\documentclass{article}

\PassOptionsToPackage{numbers, compress}{natbib}
 \usepackage[preprint]{neurips_2026}

\usepackage[utf8]{inputenc} 
\usepackage[T1]{fontenc}    
\usepackage{hyperref}       
\usepackage{url}            
\usepackage{booktabs}       
\usepackage{amsfonts}       
\usepackage{nicefrac}       
\usepackage{microtype}      
\usepackage{amsmath}
\usepackage{amssymb}
\usepackage{bm}
\usepackage{graphicx}
\usepackage{amsthm}
\usepackage{subcaption}
\usepackage{algorithmicx}
\usepackage{algpseudocode}
\usepackage{algorithm}

\usepackage{pifont}
\usepackage{diagbox}

\usepackage{wrapfig}
\usepackage{booktabs}
\usepackage{multirow}
\usepackage[table]{xcolor}
\usepackage{array}
\usepackage{makecell}

\usepackage{inconsolata}                       
\usepackage{listings}
\newfloat{listing}{htbp}{lol}
\floatname{listing}{Listing}

\newcommand{\cmark}{\ding{51}}%
\newcommand{\xmark}{\ding{55}}%

\newtheorem{definition}{Definition}[section]
\newtheorem{property}{Property}[section]

\usepackage{comment}

\title{Differentiable Efficient Operator Search}

\author{%
\textbf{Xiaohuan Pei}$^{1}$ \quad
\textbf{Jiyuan Zhang}$^{2}$ \quad
\textbf{Yuanfan Guo}$^{2}$ \quad
\textbf{Weiguo Feng}$^{2}$ \\
\textbf{Tao Huang}$^{3}$ \quad
\textbf{Cho-Jui Hsieh}$^{4}$ \quad
\textbf{Chang Xu}$^{1}$ \\[0.5em]
$^{1}$The University of Sydney \quad
$^{2}$ByteDance \\
$^{3}$Shanghai Jiao Tong University \quad
$^{4}$University of California, Los Angeles \\[0.3em]
\texttt{xiaohuan.pei@sydney.edu.au} 
}

\begin{document}

\maketitle

\begin{abstract}
Efficient models have largely relied on human-designed reduction operators, such as pruning, merging, pooling, and adaptive reweighting, which could happen at any time and anywhere. We show that these seemingly different methods can be unified as different operating regimes of a single shared operator space. Based on this observation, we introduce \textbf{Efficient Operator Search}, a unified framework in which continuous parameters control whether token information is removed, sharply merged, uniformly pooled, or softly redistributed.
Instead of hand-crafting operator compositions, we define a \emph{efficient search space}: a differentiable parameterization of layer activation, retention budget, and operator regime, and a \emph{efficient search policy}: minimizing the expectation of the task loss under one-sided budget and cost constraints. The pipeline leads to a broader shift from an \textit{\textbf{efficient design problem}} to an \textit{\textbf{operator search problem}}, suggesting a new paradigm for future efficient modeling.
Interestingly, we find that most prior mainstream baselines can be treated as special cases of this shared operator space, and further reveal consistent operator patterns across different benchmarks. Even as an early exploration of this new paradigm, our proposed method achieves competitive results on various benchmarks. We believe that it provides a unified view of prior methods, a principled lens for understanding current differences, and a general foundation for future efficient modeling. Webpage: \href{https://www.terrypei.com/eos}{\texttt{EOS}}.
\end{abstract}

\section{Introduction}
\label{sec:introduction}

\definecolor{opcolor}{RGB}{0,100,200}       
\definecolor{layercolor}{RGB}{200,50,0}     
\definecolor{budgetcolor}{RGB}{0,150,50}    

Scaling multi-modal foundation models has led to substantial performance gains across visual reasoning, multi-modal understanding, instruction following, and long-context multimodal reasoning~\citep{liu2023visual,li2023blip,team2023gemini,team2024gemini,hurst2024gpt,wang2024qwen2,li2024llava}. However, these gains often come with substantial inference cost, as dense visual tokens are repeatedly processed by large language backbones, especially in high-resolution, multi-image, and long-context scenarios~\citep{zhang2024long,liu2026global,shi2025catching,shi2026q,visionzip,prumerge,pei2025action}.
Existing work improves the efficiency of foundation models through human-designed token reduction operators, including pruning, merging, pooling, and adaptive reweighting. For example, SparseVLM~\cite{zhang2024sparsevlm} removes visually unimportant tokens according to attention scores, while ToMe~\cite{tome} merges redundant tokens to shorten the input sequence while preserving information. Although these methods demonstrate promising accuracy-efficiency trade-offs, they are usually developed as separate compression recipes with manually chosen layers, budgets, and operator forms, leaving the shared structure behind these corner operators largely underexplored.

\begin{figure}[t]
\centering
\begin{subfigure}[t]{0.42\linewidth}
\centering
\includegraphics[width=\linewidth,height=4.7cm,keepaspectratio]{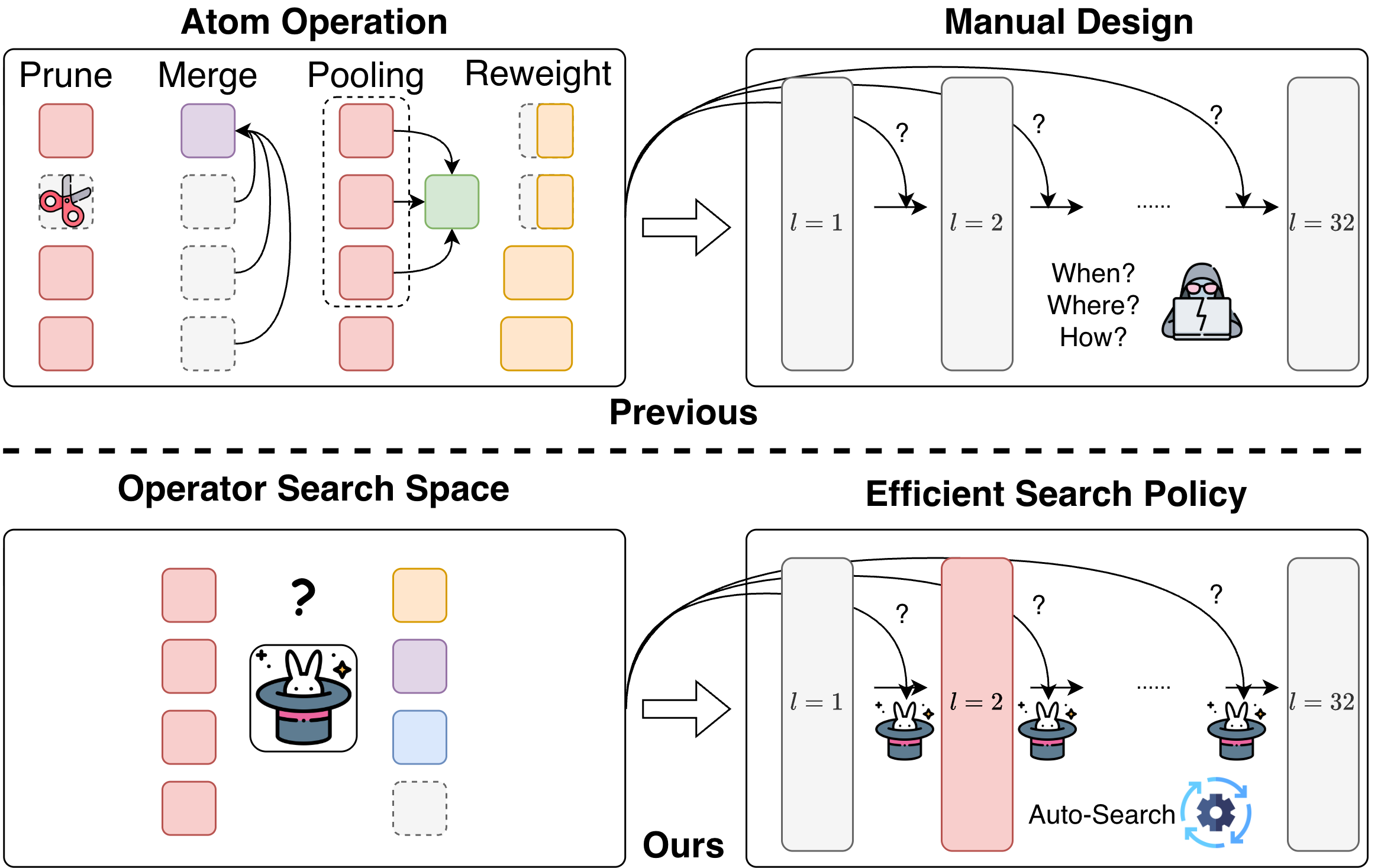}
\caption{Manual design vs. automatic search.}
\label{fig:teaser_prev_ours}
\end{subfigure}
\hfill
\begin{subfigure}[t]{0.56\linewidth}
\centering
\includegraphics[width=\linewidth,height=4.7cm,keepaspectratio]{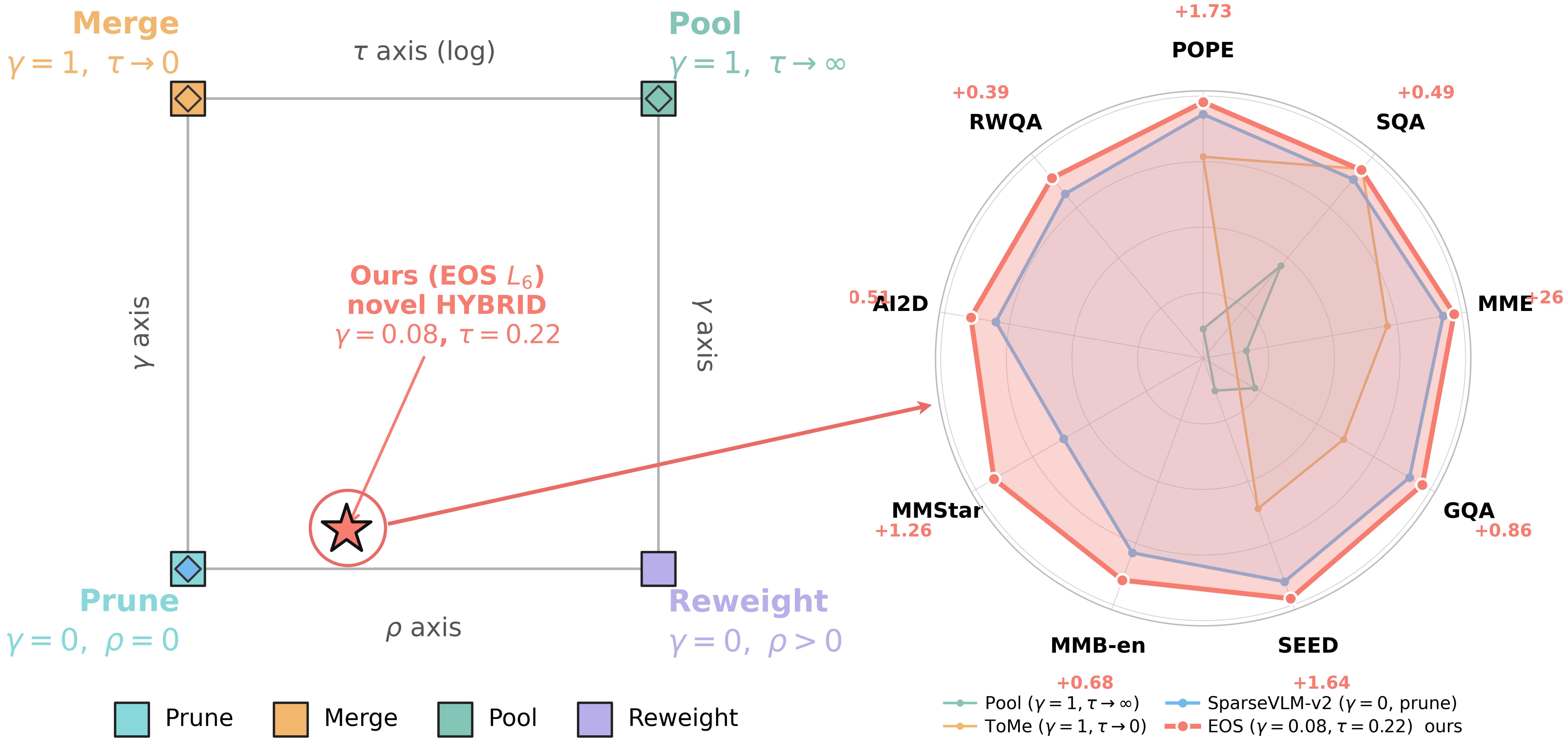}
\caption{Unified operator space and performance.}
\label{fig:teaser_intro_eos}
\end{subfigure}

\caption{
\textbf{Overview of Efficient Operator Search.}
(a) EOS replaces manually designed reduction recipes with automatic operator search.
(b) The searched hybrid operator lies inside the unified operator space and improves performance under the same token budget.
}
\vspace{-1.2em}
\label{fig:eos_overview}
\end{figure}


Despite their different implementations, existing reduction behaviors can be interpreted as different ways of discarding, transferring, or redistributing token information. This interpretation motivates a shared formulation in which representative efficient operators are controlled by continuous parameters. Specifically, an information-transfer gate $\gamma$ determines whether discarded token information is removed or preserved, while an assignment temperature $\tau$ controls whether the preserved information is sharply merged, uniformly pooled, or softly redistributed. Under this view, pure pruning, hard merging, average pooling, and adaptive reweighting naturally emerge as different operating regimes of the same operator space (Figure~\ref{fig:eos_overview}).
Based on this observation, we propose \textbf{Efficient Operator Search}, a unified framework that reframes efficient models from an \emph{operator design problem} into an \emph{operator search problem}. Instead of manually crafting another fixed compression rule, our method searches over where to compress, how many tokens to retain, and which reduction behavior to apply. This enables flexible combinations of multiple compression primitives under different accuracy-efficiency constraints.
We conduct extensive experiments on representative multi-modal benchmarks. The results show that our framework can recover strong hand-designed baselines and further discover better operator configurations, achieving improved performance-efficiency trade-offs.

Our contributions can be summarized from three aspects. (1) We introduce a unified operator space that reinterprets existing token reduction methods as different regimes of one shared formulation. (2) We propose Efficient Operator Search, which jointly searches layer selection, token budgets, and continuous operator parameters instead of relying on manually designed recipes. (3) We demonstrate the proposed framework can recover existing baselines and discover stronger hybrid configurations for efficient multimodal foundation models.

\section{Method}
\label{sec:method}

\begin{figure*}[t]
\centering
\includegraphics[width=\linewidth]{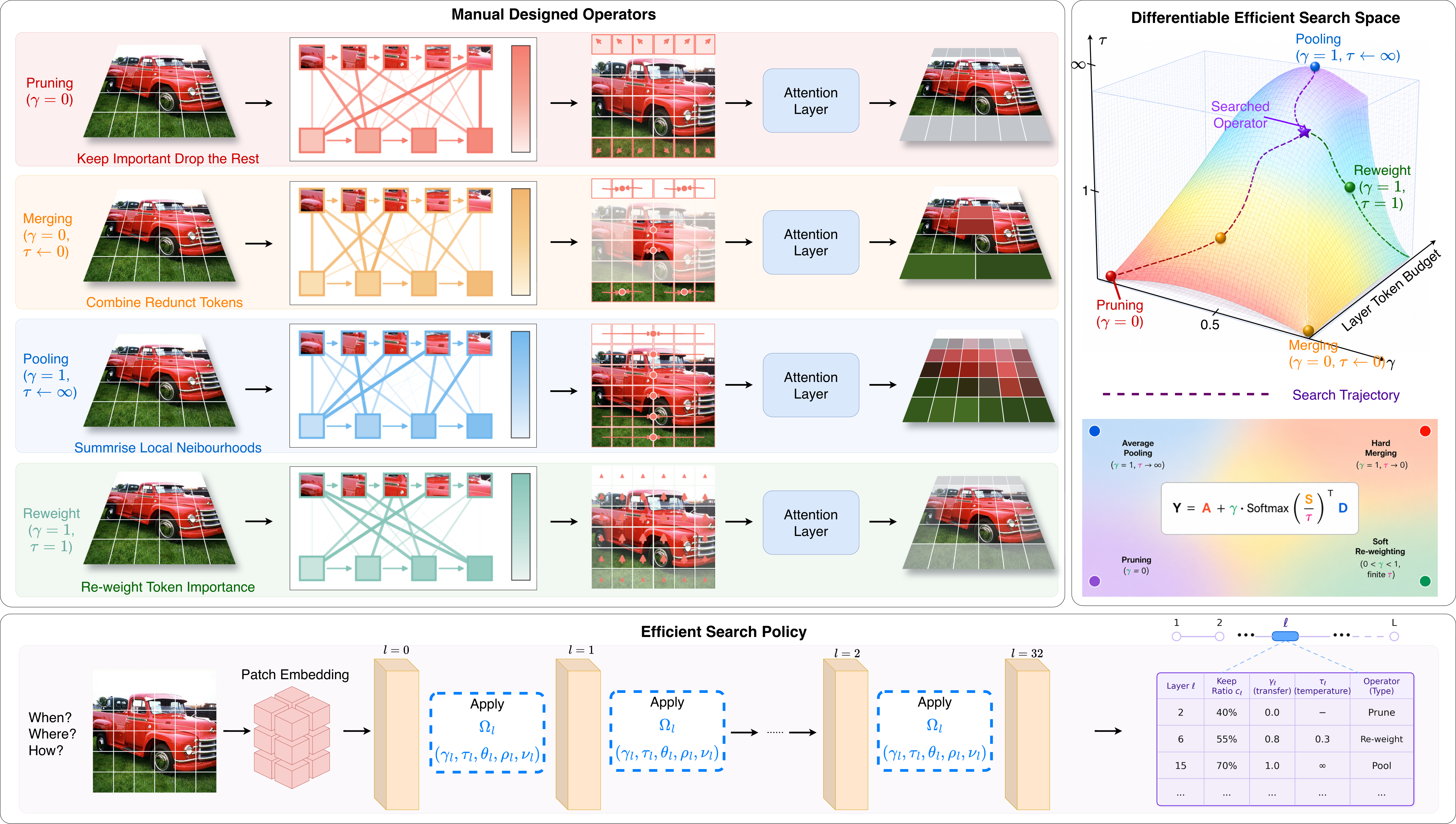}
\caption{
\textbf{Overview of Efficient Operator Search.}
Given a frozen multimodal foundation model, EOS parameterizes token reduction at each decoder layer by three coupled components: layer activation ${\color{layercolor}g_l}$, retention budget ${\color{budgetcolor}c_l}$, and operator regime ${\color{opcolor}\Omega_l=(\gamma_l,\tau_l,\theta_l,\rho_l,\nu_l)}$.
At each active layer, important visual tokens are retained as anchors, while the remaining candidates are processed by a unified reduction operator.
Depending on the learned operator parameters, discarded information can be removed, sharply merged, uniformly pooled, or softly redistributed to the anchors.
}
\vspace{-1.2em}
\label{fig:main}
\end{figure*}

We aim to transform efficient foundation models from manually designed compression recipes into a searchable operator space.
Given a frozen multimodal foundation model with $L$ decoder layers, existing token-reduction methods typically hand-specify \emph{where} to reduce visual tokens, \emph{how many} tokens to retain, and \emph{which} operator to apply.
In contrast, our \textbf{Efficient Operator Search} (EOS) parameterizes these choices as learnable search variables and jointly optimizes them with differentiable relaxations under task and efficiency constraints.
As shown in Figure~\ref{fig:main}, EOS selects important visual-token anchors at each active reduction layer, then applies a unified operator to decide whether the remaining token information is discarded, transferred, pooled, or reweighted.

\subsection{Preliminary}
\label{subsec:preliminary}

Let $\mathbf{X}^{(l)} \in \mathbb{R}^{N_l \times d}$ denote the visual hidden states at decoder layer $l$, where $N_l$ is the number of visual tokens and $d$ is the hidden dimension.
A token-reduction operation partitions $\mathbf{X}^{(l)}$ into a retained anchor set and a reduction-candidate set:
\begin{equation}
    \mathbf{X}^{(l)} = \mathbf{A}^{(l)} \cup \mathbf{D}^{(l)},
    \qquad
    \mathbf{A}^{(l)} \in \mathbb{R}^{K_l \times d},
    \quad
    \mathbf{D}^{(l)} \in \mathbb{R}^{M_l \times d},
    \label{eq:partition}
\end{equation}
where $K_l$ is the number of retained anchors, $M_l = N_l - K_l$ is the number of reduction candidates, and $\mathbf{A}^{(l)} \cap \mathbf{D}^{(l)} = \emptyset$.
The anchors are selected according to an importance score $\mathbf{q}^{(l)} \in \mathbb{R}^{N_l}$, which can be instantiated by attention-based or similarity-based token importance following prior efficient inference methods.
The remaining question is how to process $\mathbf{D}^{(l)}$ after anchor selection.
Pure pruning directly removes $\mathbf{D}^{(l)}$, token merging transfers each discarded token to a similar anchor, pooling distributes discarded information uniformly, and adaptive reweighting rescales anchors according to discarded-token importance.
EOS treats these behaviors as different operating regimes of one shared operator family.

\subsection{Unified Reduction Operator}
\label{subsec:unified_operator}

For simplicity, we omit the layer index $l$ when the context is clear.
Given anchors $\mathbf{A} \in \mathbb{R}^{K \times d}$ and reduction candidates $\mathbf{D} \in \mathbb{R}^{M \times d}$, we first compute the normalized similarity matrix:
\begin{equation}
    \mathbf{S}
    =
    \bar{\mathbf{D}} \bar{\mathbf{A}}^\top
    \in \mathbb{R}^{M \times K},
    \qquad
    \bar{\mathbf{x}}_i
    =
    \frac{\mathbf{x}_i}{\|\mathbf{x}_i\|_2+\epsilon},
    \label{eq:similarity}
\end{equation}
where $\mathbf{x}_i$ denotes a row token feature from either $\mathbf{A}$ or $\mathbf{D}$, the normalization is applied token-wise, and $\epsilon>0$ is a small constant used for numerical stability.
Each entry $S_{ij}$ measures the cosine similarity between reduction candidate $\mathbf{d}_i$ and anchor token $\mathbf{a}_j$.
The assignment from reduction candidates to anchors is given by
\begin{equation}
    \mathbf{W}
    =
    \mathrm{softmax}_{\mathrm{anchor}}
    \left(
    \frac{\mathbf{S}}{\textcolor{opcolor}{\tau_l}}
    \right)
    \in \mathbb{R}^{M \times K},
    \label{eq:assignment}
\end{equation}
where $\mathrm{softmax}_{\mathrm{anchor}}(\cdot)$ normalizes each row over the anchor dimension, and $\textcolor{opcolor}{\tau_l} > 0$ is an assignment temperature.
A smaller $\textcolor{opcolor}{\tau_l}$ induces sharper nearest-anchor assignment, while a larger $\textcolor{opcolor}{\tau_l}$ yields smoother assignment across anchors.

We further define a per-token transfer gate:
\begin{equation}
    m_i
    =
    \sigma
    \left(
    \beta(\max_j S_{ij} - \textcolor{opcolor}{\theta_l})
    \right),
    \qquad
    \mathbf{m} = [m_1,\ldots,m_M]^\top,
    \label{eq:transfer_gate}
\end{equation}
where $\textcolor{opcolor}{\theta_l}$ is a similarity threshold and $\beta$ controls the gate sharpness.
The unified information-transfer operation is then given by
\begin{equation}
    \tilde{\mathbf{A}}
    =
    \mathbf{A}
    +
    \textcolor{opcolor}{\gamma_l}\,
    \mathbf{W}^{\top}(\mathbf{D}\odot \mathbf{m}),
    \label{eq:unified_transfer}
\end{equation}
where $\mathbf{m}$ is broadcast along the feature dimension, and $\textcolor{opcolor}{\gamma_l} \in [0,1]$ is an information-transfer gate.
When $\textcolor{opcolor}{\gamma_l}=0$, the reduction candidates have no explicit feature transfer and are effectively pruned.
When $\textcolor{opcolor}{\gamma_l}>0$, their information is transferred to the retained anchors according to the assignment matrix $\mathbf{W}$.

To include adaptive reweighting and stabilize the output scale, we further apply
\begin{equation}
    \hat{\mathbf{A}}
    =
    \tilde{\mathbf{A}}
    \odot
    \left(1 + \textcolor{opcolor}{\rho_l}\,\mathbf{s}_{\mathrm{imp}}\right),
    \qquad
    \hat{\mathbf{A}}
    \leftarrow
    (1-\textcolor{opcolor}{\nu_l})\hat{\mathbf{A}}
    +
    \textcolor{opcolor}{\nu_l}
    \frac{\|\mathbf{A}\|_F}{\|\hat{\mathbf{A}}\|_F+\epsilon}
    \hat{\mathbf{A}},
    \label{eq:reweight_norm}
\end{equation}
where $\textcolor{opcolor}{\rho_l}$ denotes the anchor-reweighting strength, $\textcolor{opcolor}{\nu_l}$ denotes the norm-preservation coefficient, and $\mathbf{s}_{\mathrm{imp}} \in \mathbb{R}^{K}$ denotes the anchor-level importance induced by reduction candidates.
Here, $\textcolor{opcolor}{\gamma_l}$ controls explicit feature transfer, while $\textcolor{opcolor}{\rho_l}$ controls indirect anchor rescaling.
In practice, $\mathbf{s}_{\mathrm{imp}}$ is given by
\begin{equation}
    \mathbf{s}_{\mathrm{imp}}
    =
    \mathrm{softmax}_{\mathrm{anchor}}(\mathbf{S})^\top
    \mathrm{softmax}_{\mathrm{candidate}}(\mathbf{q}_{\mathcal{D}}),
    \label{eq:anchor_importance}
\end{equation}
where $\mathbf{q}_{\mathcal{D}}$ denotes the importance scores of the reduction candidates, and $\mathrm{softmax}_{\mathrm{candidate}}(\cdot)$ normalizes over the reduction-candidate dimension.

\begin{definition}[Unified efficient operator]
\label{def:unified_operator}
At layer $l$, an efficient reduction operator is defined by the parameter tuple
\begin{equation}
    \textcolor{opcolor}{\Omega_l}
    =
    \left(
    \textcolor{opcolor}{\gamma_l},
    \textcolor{opcolor}{\tau_l},
    \textcolor{opcolor}{\theta_l},
    \textcolor{opcolor}{\rho_l},
    \textcolor{opcolor}{\nu_l}
    \right),
    \label{eq:operator_tuple}
\end{equation}
and maps the visual hidden states from $\mathbf{X}^{(l)} \in \mathbb{R}^{N_l \times d}$ to a reduced representation $\hat{\mathbf{A}}^{(l)} \in \mathbb{R}^{K_l \times d}$ through Eqs.~\ref{eq:similarity}--\ref{eq:reweight_norm}.
The parameters $\textcolor{opcolor}{\gamma_l}$ and $\textcolor{opcolor}{\tau_l}$ determine the dominant reduction regime, while $\textcolor{opcolor}{\theta_l}$, $\textcolor{opcolor}{\rho_l}$, and $\textcolor{opcolor}{\nu_l}$ provide token-level gating, anchor reweighting, and scale preservation.
\end{definition}

The parameter tuple $\textcolor{opcolor}{\Omega_l}$ defines the reduction behavior at a single layer.
EOS further couples this operator regime with layer activation and token budget, yielding a searchable space over where to compress, how much to compress, and how to process the reduced tokens.

\subsection{Efficient Search Space}
\label{subsec:search_space}

EOS searches over three coupled dimensions: \textcolor{layercolor}{layer activation}, \textcolor{budgetcolor}{token budget}, and \textcolor{opcolor}{operator regime}.
For all decoder layers, the complete search space is given by
\begin{equation}
\boxed{
\boldsymbol{\Theta}
=
\Bigl\{
\underbrace{\textcolor{layercolor}{g_l}}_{\text{which layers}},
\;
\underbrace{\textcolor{budgetcolor}{c_l}}_{\text{how much}},
\;
\underbrace{\textcolor{opcolor}{\gamma_l,\tau_l,\theta_l,\rho_l,\nu_l}}_{\text{which operator}}
\Bigr\}_{l=0}^{L-1}
}
\label{eq:search_space}
\end{equation}
where $\textcolor{layercolor}{g_l}$ determines whether layer $l$ applies token reduction, $\textcolor{budgetcolor}{c_l}$ controls the fraction of visual tokens to discard, and $\textcolor{opcolor}{(\gamma_l,\tau_l,\theta_l,\rho_l,\nu_l)}$ specify the operator regime.
All variables are represented by unconstrained learnable parameters and mapped to valid ranges through differentiable transformations:
\begin{equation}
\begin{aligned}
    \textcolor{layercolor}{g_l} &= \sigma(w_l^g \cdot T_g),
    &
    \textcolor{budgetcolor}{c_l} &= \sigma(w_l^c \cdot T_c),
    &
    \textcolor{opcolor}{\gamma_l} &= \sigma(w_l^\gamma), \\
    \textcolor{opcolor}{\tau_l} &= \mathrm{softplus}(w_l^\tau)+\epsilon,
    &
    \textcolor{opcolor}{\theta_l} &= 2\sigma(w_l^\theta)-1,
    &
    \textcolor{opcolor}{\rho_l} &= \sigma(w_l^\rho),
    \qquad
    \textcolor{opcolor}{\nu_l} = \sigma(w_l^\nu),
\end{aligned}
\label{eq:param_activation}
\end{equation}
where $T_g$ and $T_c$ are optional scaling temperatures used to sharpen the structural gates, and the same numerical constant $\epsilon>0$ ensures a strictly positive assignment temperature.
The effective number of retained visual tokens at layer $l$ is given by
\begin{equation}
    K_l
    =
    \max
    \left(
    K_{\min},
    \left\lfloor
    \bigl(1-\textcolor{layercolor}{g_l}\textcolor{budgetcolor}{c_l}\bigr)N_l
    \right\rfloor
    \right).
    \label{eq:retain_number}
\end{equation}
When $\textcolor{layercolor}{g_l}$ is close to zero, the layer is effectively inactive and the visual sequence is passed to the next layer unchanged.
When $\textcolor{layercolor}{g_l}$ is active, $\textcolor{budgetcolor}{c_l}$ determines the compression strength, and the operator tuple $\textcolor{opcolor}{\Omega_l}$ determines how reduction candidates influence retained anchors.
Although Eq.~\ref{eq:retain_number} defines the discrete forward retain number, EOS optimizes its underlying variables through the relaxation introduced in Sec.~\ref{subsec:differentiable_selection}.
The structural variables $(\textcolor{layercolor}{g_l},\textcolor{budgetcolor}{c_l})$ answer \emph{where} and \emph{how much} to compress, while the operator variables $\textcolor{opcolor}{(\gamma_l,\tau_l,\theta_l,\rho_l,\nu_l)}$ answer \emph{which reduction behavior} to apply.
Such factorization enables EOS to recover existing hand-designed methods by fixing part of $\boldsymbol{\Theta}$, while also allowing new hybrid configurations to be discovered through continuous optimization.

\subsection{Differentiable Token Selection}
\label{subsec:differentiable_selection}

Since top-$K_l$ anchor selection is discrete, we use a straight-through soft boundary relaxation to preserve the forward behavior while allowing gradients to update the retention budget.
Let $q_{(K_l)}$ denote the detached $K_l$-th largest importance score, and define
\begin{equation}
    p_i^{\mathrm{keep}}
    =
    \sigma
    \left(
    \alpha(q_i-q_{(K_l)})
    \right)
    \cdot
    \frac{1-\textcolor{layercolor}{g_l}\textcolor{budgetcolor}{c_l}}
    {1-\mathrm{sg}(\textcolor{layercolor}{g_l}\textcolor{budgetcolor}{c_l})},
    \label{eq:soft_keep}
\end{equation}
where $\alpha$ controls the boundary sharpness and $\mathrm{sg}(\cdot)$ denotes the stop-gradient operation.
The first term approximates whether token $i$ lies above the selection threshold, while the second term provides a straight-through gradient path from $\textcolor{layercolor}{g_l}$ and $\textcolor{budgetcolor}{c_l}$ without changing the forward value.
The relaxed keep probabilities estimate the differentiable token count, while inference still uses hard top-$K$ anchors and reduction candidates in Eq.~\ref{eq:unified_transfer}.

\subsection{Differentiable Search Policy}
\label{subsec:search_policy}

The backbone multimodal foundation model is kept frozen, and only the search parameters $\boldsymbol{\Theta}$ are optimized.
Given an input-output pair $(\mathbf{x},\mathbf{y})$, EOS minimizes the task loss under one-sided efficiency constraints:
\begin{equation}
\boxed{
\min_{\boldsymbol{\Theta}}
\quad
\underbrace{\mathcal{L}_{\mathrm{task}}(\boldsymbol{\Theta})}_{\text{task loss}}
+
\underbrace{
\lambda_b
\left[
\frac{N_{\mathrm{final}}(\boldsymbol{\Theta})}{B}-1
\right]_+^2
}_{\text{token budget: } N_{\mathrm{final}} \leq B}
+
\underbrace{
\lambda_c
\left[
\sum_{l=0}^{L-1}\textcolor{layercolor}{g_l} - C
\right]_+^2
}_{\text{inference cost: active layers} \leq C}
}
\label{eq:search_objective}
\end{equation}
where $[\cdot]_+=\max(0,\cdot)$, $B$ is the target final token budget, and $C$ is the target maximum number of active reduction layers.
Here, $N_{\mathrm{final}}(\boldsymbol{\Theta})$ is estimated by 
$\bar{N}_0=N_0$, 
$\bar{N}_{l+1}=(1-\textcolor{layercolor}{g_l}\textcolor{budgetcolor}{c_l})\bar{N}_l$, 
and $N_{\mathrm{final}}(\boldsymbol{\Theta})=\bar{N}_L$.
At deployment, it is replaced by the corresponding discrete retain count after top-$K$ selection.
The first regularizer penalizes configurations that exceed the token budget, while the second penalizes configurations that activate too many reduction layers.
Both constraints are one-sided: no penalty is applied once the configuration satisfies the desired budget.
This design allows the search to prioritize task performance within a feasible efficiency region, rather than forcing unnecessary compression when the budget has already been met. The task loss is given by
\begin{equation}
\boxed{
\mathcal{L}_{\mathrm{task}}(\boldsymbol{\Theta})
=
\underbrace{
\mathcal{L}_{\mathrm{CE}}
\left(
\mathbf{y},
f_{\mathrm{MFM}}(\mathbf{x};\boldsymbol{\Theta})
\right)
}_{\text{text prediction}}
+
\textcolor{opcolor}{\lambda_a}
\underbrace{
\frac{1}{|\mathcal{A}|}
\sum_{l\in\mathcal{A}}
\left\|
\mathbf{H}^{\mathrm{red}}_l[\mathrm{text}]
-
\mathrm{sg}
\left(
\mathbf{H}^{\mathrm{full}}_l[\mathrm{text}]
\right)
\right\|_2^2
}_{\mathcal{L}_{\mathrm{align}}:\ \text{hidden-state alignment}}
}
\label{eq:task_loss}
\end{equation}
where $\mathcal{L}_{\mathrm{CE}}$ is the standard language-modeling loss, $\mathbf{H}^{\mathrm{red}}_l$ denotes the hidden states produced by the reduced forward pass, and $\mathbf{H}^{\mathrm{full}}_l$ is obtained from a reference forward pass without token reduction.
The operator $\mathrm{sg}(\cdot)$ blocks gradients through the full-model reference states, so the alignment term only updates the search parameters through the reduced forward pass.
The set $\mathcal{A}$ contains selected decoder layers for measuring the deviation between reduced and full computation.
\begin{wraptable}{r}{0.4\linewidth}
\vspace{-1.2em}
\centering
\small
\caption{One-sided constraints.}
\label{tab:onesided_constraints}
\resizebox{\linewidth}{!}{
\begin{tabular}{@{}lll@{}}
\toprule
\textbf{Constraint} & \textbf{Active} & \textbf{Inactive} \\
\midrule
Token budget 
& $N_{\mathrm{final}}>B$ 
& $N_{\mathrm{final}}\leq B$ \\
Active layers 
& $\sum_l {\color{layercolor}g_l}>C$ 
& $\sum_l {\color{layercolor}g_l}\leq C$ \\
\bottomrule
\end{tabular}
}
\vspace{-1.2em}
\end{wraptable}
The hidden-state alignment term is important because cross-entropy mainly evaluates the final textual prediction and may be insensitive to whether discarded visual information is removed, transferred, or softly redistributed.
The constraints in Eq.~\ref{eq:search_objective} are one-sided: they penalize only budget violations and vanish once the target efficiency region is reached, allowing the search to focus on preserving task accuracy.

\subsection{Optimization and Deployment}
\label{subsec:optimization_deployment}

We initialize the search from a known efficient configuration by activating a small number of reduction layers and setting $\textcolor{opcolor}{\gamma_l}$ close to zero, corresponding to a pruning-like regime.
This warm start provides a stable accuracy-efficiency point, after which gradient descent adjusts layer gates, token budgets, and operator parameters under Eq.~\ref{eq:search_objective}.

After optimization, the learned variables are converted into a deterministic inference configuration: layers with large $\textcolor{layercolor}{g_l}$ become active reduction layers, $\textcolor{budgetcolor}{c_l}$ determines the retain schedule, and the operator parameters define the final reduction behavior.
The resulting configuration $\boldsymbol{\Theta}^{\star}$ is fixed during evaluation.
For different retain budgets, we keep the learned active layers and operator regime fixed, and only rescale the final retain schedule.
Thus, EOS does not require retraining a separate reducer for every budget, but learns a general operator configuration deployable under multiple efficiency settings.

\subsection{Corner Operators as Special Cases}
\label{subsec:corner_operators}

After defining the search space and optimization policy, we now discuss its relation to existing hand-designed reduction operators.
The key property of EOS is that its operator regime $\textcolor{opcolor}{\Omega_l}$ admits several standard operators as limiting cases.
This establishes that pruning, merging, pooling, and adaptive reweighting are not independent mechanisms, but different operating regimes of the same parameterized operator family.

\vspace{-0.3em}
\begin{definition}[Pure pruning]
\label{def:pure_pruning}
Pure pruning is recovered by setting
\begin{equation}
    \textcolor{opcolor}{\gamma_l}=0,\qquad
    \textcolor{opcolor}{\rho_l}=0.
    \label{eq:prune_setting}
\end{equation}
In this case, Eq.~\ref{eq:unified_transfer} reduces to
\begin{equation}
    \tilde{\mathbf{A}}
    =
    \mathbf{A}
    +
    0 \cdot \mathbf{W}^{\top}(\mathbf{D}\odot \mathbf{m})
    =
    \mathbf{A}.
    \label{eq:prune_recover}
\end{equation}
Thus, the reduction candidates do not contribute to the retained representation and are directly removed from subsequent computation.
\end{definition}

\vspace{-0.5em}
\begin{definition}[Hard token merging]
\label{def:hard_merging}
Hard token merging is recovered by setting
\begin{equation}
    \textcolor{opcolor}{\gamma_l}=1,\quad
    \textcolor{opcolor}{\tau_l}\rightarrow 0^+,\quad
    \textcolor{opcolor}{\theta_l}\rightarrow -1,\quad
    \textcolor{opcolor}{\rho_l}=0,\quad
    \textcolor{opcolor}{\nu_l}=0.
    \label{eq:merge_setting}
\end{equation}
The setting $\textcolor{opcolor}{\theta_l}\rightarrow -1$ makes $m_i\rightarrow 1$, disabling the transfer gate.
Since $\mathbf{W}=\mathrm{softmax}_{\mathrm{anchor}}(\mathbf{S}/\textcolor{opcolor}{\tau_l})$, the limiting case $\textcolor{opcolor}{\tau_l}\rightarrow 0^+$ yields
\begin{equation}
    W_{ij}
    \rightarrow
    \mathbf{1}\!\left[
    j=\arg\max_k S_{ik}
    \right],
    \qquad
    \hat{\mathbf{a}}_j
    =
    \mathbf{a}_j
    +
    \sum_{i:\; j=\arg\max_k S_{ik}}
    \mathbf{d}_i.
    \label{eq:merge_recover}
\end{equation}
Therefore, each reduction candidate is transferred to its nearest anchor, which corresponds to a hard nearest-neighbor additive merge.
\end{definition}

\vspace{-0.5em}
\begin{definition}[Uniform pooling]
\label{def:uniform_pooling}
Uniform pooling is recovered by setting
\begin{equation}
    \textcolor{opcolor}{\gamma_l}=1,\quad
    \textcolor{opcolor}{\tau_l}\rightarrow \infty,\quad
    \textcolor{opcolor}{\theta_l}\rightarrow -1,\quad
    \textcolor{opcolor}{\rho_l}=0,\quad
    \textcolor{opcolor}{\nu_l}=0.
    \label{eq:pool_setting}
\end{equation}
The setting $\textcolor{opcolor}{\theta_l}\rightarrow -1$ makes $m_i\rightarrow 1$, disabling the transfer gate.
In this limiting case, the assignment distribution becomes uniform:
\begin{equation}
    W_{ij}
    =
    \frac{\exp(S_{ij}/\textcolor{opcolor}{\tau_l})}
    {\sum_{k=1}^{K}\exp(S_{ik}/\textcolor{opcolor}{\tau_l})}
    \rightarrow
    \frac{1}{K},
    \qquad
    \hat{\mathbf{a}}_j
    =
    \mathbf{a}_j
    +
    \frac{1}{K}
    \sum_{i=1}^{M}
    \mathbf{d}_i.
    \label{eq:pool_recover}
\end{equation}
Therefore, every anchor receives the same averaged contribution from the reduction candidates, corresponding to uniformly pooling discarded information into the retained anchors.
\end{definition}

\vspace{-0.5em}
\begin{definition}[Adaptive reweighting]
\label{def:adaptive_reweighting}
Adaptive reweighting is recovered by setting
\begin{equation}
    \textcolor{opcolor}{\gamma_l}=0,\qquad
    \textcolor{opcolor}{\rho_l}>0,\qquad
    \textcolor{opcolor}{\nu_l}=0.
    \label{eq:reweight_setting}
\end{equation}
Since $\textcolor{opcolor}{\gamma_l}=0$, no reduction candidate is explicitly transferred to the anchors through Eq.~\ref{eq:unified_transfer}.
The output is instead given by the reweighting term in Eq.~\ref{eq:reweight_norm}:
\begin{equation}
    \hat{\mathbf{a}}_j
    =
    \left(1+\textcolor{opcolor}{\rho_l}\,s_{\mathrm{imp},j}\right)\mathbf{a}_j.
    \label{eq:reweight_recover}
\end{equation}
Thus, reduction candidates affect the retained representation only through anchor-level importance rescaling, without explicit token transfer.
\end{definition}

\vspace{-0.5em}
\begin{property}[Operator unification]
\label{prop:operator_unification}
The unified operator in Definition~\ref{def:unified_operator} admits pure pruning, hard merging, uniform pooling, and adaptive reweighting as special cases under different parameter settings of $\textcolor{opcolor}{\Omega_l}$.
\end{property}

\begin{table}[t]
\centering
\caption{
\textbf{Prior methods as fixed points in the EOS search space.}
EOS optimizes these dimensions jointly, allowing both corner operators and interior hybrid operators to be discovered.
}
\label{tab:unification}
\resizebox{\textwidth}{!}{%
\begin{tabular}{@{}l|ccc|ccccc|c@{}}
\toprule
\multirow{2}{*}{\textbf{Method}} &
\multicolumn{3}{c|}{\textbf{Structure}} &
\multicolumn{5}{c|}{\textbf{Operator Regime}} &
\multirow{2}{*}{\textbf{Searchable}} \\
& \textcolor{layercolor}{Layers $\mathcal{R}$}
& \textcolor{budgetcolor}{Budget $c_l$}
& \textcolor{layercolor}{Gate $g_l$}
& \textcolor{opcolor}{$\gamma$}
& \textcolor{opcolor}{$\tau$}
& \textcolor{opcolor}{$\theta$}
& \textcolor{opcolor}{$\rho$}
& \textcolor{opcolor}{$\nu$}
& \\
\midrule
FastV~\cite{fastv}
& fixed & fixed & fixed & $0$ & -- & -- & $0$ & -- & \xmark \\
ToMe~\cite{tome}
& fixed/all & uniform & fixed & $1$ & $\to 0$ & $\to -1$ & $0$ & $0$ & \xmark \\
SparseVLM~\cite{zhang2024sparsevlm}
& fixed & schedule & fixed & $0$ & -- & -- & $0$ & -- & \xmark \\
VisionZip~\cite{visionzip}
& fixed & fixed & fixed & $0$ & -- & -- & $0$ & -- & \xmark \\
PruMerge+~\cite{prumerge}
& fixed & fixed & fixed & mixed & manual & -- & $0$ & -- & \xmark \\
\midrule
\rowcolor{blue!8}
\textbf{EOS}
& learned & learned & learned
& learned & learned & learned & learned & learned
& \cmark \\
\bottomrule
\end{tabular}%
}
\vspace{-1.2em}
\end{table}

The four corner operators arise from different regimes of $\textcolor{opcolor}{\Omega_l}$, and Table~\ref{tab:unification} places representative baselines into the proposed search space.
Existing methods can be interpreted as manually specified points in $\boldsymbol{\Theta}$, since they typically fix the active layers, token budgets, or operator regimes by design.
EOS instead makes these dimensions searchable, thereby generalizing representative reduction behaviors and enabling intermediate hybrid operators beyond the manually selected corners.


\begin{table}[t]
\centering
\caption{\textbf{Empirical verification.}
Our unified operator reproduces existing token-reduction methods under the same setting, and further improves performance after search.}
\label{tab:reproduce}
\resizebox{\textwidth}{!}{%
\begin{tabular}{@{} l l | c c c c c | l @{}}
\toprule
\textbf{Method} & \textbf{Search-space setting} & \textbf{POPE} & \textbf{SQA} & \textbf{MME} & \textbf{GQA} & \textbf{TextVQA} & \textbf{Match?} \\
\midrule
\multicolumn{8}{@{}l}{\emph{Original code implementations:}} \\
\midrule
SparseVLM+ (original code)
& ---
& 85.79 & 64.80 & 1869 & 60.95 & 45.25
& --- \\
SparseVLM v1 (original code)
& ---
& 85.22 & 64.15 & 1825 & 59.45 & 44.17
& --- \\
\midrule
\multicolumn{8}{@{}l}{\emph{Special cases of our unified operator:}} \\
\midrule
\rowcolor{red!5}
Ours $\equiv$ SparseVLM+
& $\textcolor{opcolor}{\gamma{=}0},\;\textcolor{layercolor}{\mathcal{R}{=}\{2,6,15\}},\;\textcolor{budgetcolor}{c_l{=}\text{V2 sched.}}$
& 85.96 & 65.20 & 1853 & 60.13 & 43.86
& $\approx$ \ding{51} \\
\rowcolor{red!5}
Ours $\equiv$ SparseVLM v1
& $\textcolor{opcolor}{\gamma{=}0},\;\textcolor{layercolor}{\mathcal{R}{=}\{2,6,15\}},\;\textcolor{budgetcolor}{c_l{=}\text{V1 sched.}}$
& 85.22 & 64.15 & 1825 & 59.45 & 44.17
& \textbf{Exact} \ding{51} \\
\midrule
\multicolumn{8}{@{}l}{\emph{Searched operators in our unified space:}} \\
\midrule
\rowcolor{green!8}
Ours (searched, layers fixed)
& $\textcolor{opcolor}{\gamma_6{=}0.08,\;\tau_6{=}0.22},\;\text{rest }\textcolor{opcolor}{\gamma{=}0}$
& 85.61 & \textbf{65.05} & 1867 & 60.81 & 44.87
& $\approx$ V2 \\
\rowcolor{green!8}
Ours (searched)
& $\textcolor{opcolor}{\gamma_6{=}0.08,\;\tau_6{=}0.22}$
& \textbf{85.78} & 64.85 & \textbf{1876} & \textbf{60.92} & \textbf{45.29}
& \textbf{Beats V2} \ding{51} \\
\bottomrule
\end{tabular}%
}
\vspace{-1.3em}
\end{table}

\begin{figure*}[t]
\centering
\includegraphics[width=\linewidth]{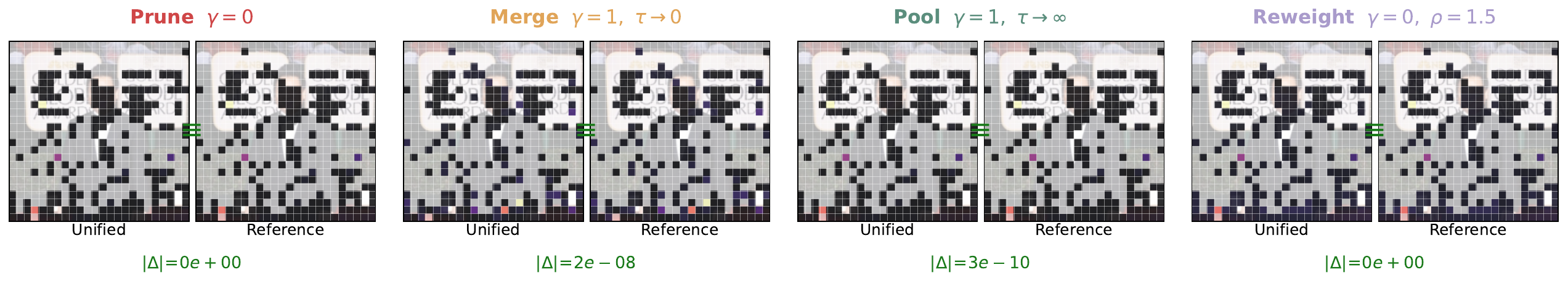}
\vspace{-0.5em}
\caption{
\textbf{Numerical verification of corner operators.}
Our unified operator reproduces \textsc{Prune}, \textsc{Merge}, \textsc{Pool}, and \textsc{Reweight} under their corresponding settings in $\textcolor{opcolor}{\Omega_l}$. 
}
\vspace{-0.5em}
\label{fig:operator_equivalence}
\end{figure*}

\section{Experiments}
\label{sec:experiments}

\subsection{Experimental Setup}
\label{subsec:experimental_setup}

\textbf{Model and baselines.}
We evaluate EOS on frozen LLaVA~\cite{liu2023visual} and compare it with representative corner operators, including pruning-based \textsc{SparseVLM-v1/v2}~\cite{zhang2024sparsevlm}, merging-based \textsc{ToMe}~\cite{tome}, and pooling-based \textsc{Pool}.
All methods use the same retained visual-token budgets for fair comparison.

\textbf{Benchmarks.}
We evaluate on representative multimodal benchmarks.
Following previous baselines, the prior benchmark group includes POPE~\cite{pope}, SQA~\cite{sqa}, MME\cite{mme}, GQA~\cite{gqa}, TextVQA\cite{textvqa}, and SEED\cite{seedbench}.
To provide a broader evaluation, the extended benchmark group includes MMStar\cite{mmstar}, RealWorldQA (RWQA), AI2D\cite{ai2d}, OCRBench\cite{liu2024ocrbench}, ChartQA\cite{chartqa}, and MMBench-en\cite{mmbench}.

\textbf{Evaluation protocol.}
We report results  across four retained-token budgets: $r\in\{192,128,64,16\}$, corresponding to progressively stronger visual-token reduction.
For \textsc{ToMe} and \textsc{Pool}, we use the same reducer-layer placement $\mathcal{R}=\{2,6,15\}$ for controlled comparison with pruning-based baselines.
EOS uses the searched operator configuration $\boldsymbol{\Theta}^{\star}$ and keeps it fixed during evaluation.

\subsection{Main Results}
\label{subsec:main_results}

Table~\ref{tab:main_retainsweep} presents the main comparison across different retained-token budgets.
EOS consistently achieves competitive or superior performance across both prior and extended benchmarks.
The advantage becomes more pronounced under aggressive token reduction, indicating that the searched hybrid operator is more robust than manually selected corner operators when the visual-token budget becomes highly constrained.

\begin{table*}[t]
\centering
\caption{
\textbf{Main results across retained-token budgets.}
EOS is compared with pruning-, merging-, and pooling-based baselines under identical visual-token budgets.
}
\label{tab:main_retainsweep}
\resizebox{\textwidth}{!}{%
\begin{tabular}{@{} l | c c c c c c | c c c c c c || c @{}}
\toprule
\multirow{2}{*}{\textbf{Method}}
& \multicolumn{6}{c|}{\textbf{Prior Benchmarks}}
& \multicolumn{6}{c||}{\textbf{Extended Benchmarks}}
& \multirow{2}{*}{\textbf{Overall Avg.}} \\
\cmidrule(lr){2-7}\cmidrule(lr){8-13}
& POPE & SQA & MME & GQA & TextVQA & SEED
& MMStar & RWQA & AI2D & OCRBench & ChartQA & MMB-en & \\
\midrule

\multicolumn{14}{@{}l}{\emph{Retain = 192 visual tokens}} \\
\midrule
SparseVLM-v1
& 85.22 & 64.15 & 1825 & 59.41 & 44.24 & 64.26
& 34.97 & 53.59 & 54.73 & 30.60 & 17.52 & 63.92
& 53.15 \\
SparseVLM-v2
& \textbf{85.79} & 64.80 & 1869 & \textbf{60.95} & 45.25 & 65.59
& 33.98 & 54.77 & 55.38 & 30.40 & 18.12 & 63.75
& 53.79 \\
ToMe $\mathcal{R}{=}\{2,6,15\}$
& 85.09 & 64.45 & 1858 & 59.64 & 43.25 & 64.62
& 33.77 & 53.73 & 55.18 & 28.90 & 16.96 & 63.40
& 52.95 \\
Pool $\mathcal{R}{=}\{2,6,15\}$
& 52.38 & 60.54 & 1287 & 43.57 & 8.17 & 39.06
& 24.64 & 43.92 & 50.36 & 2.40 & 11.80 & 24.14
& 33.91 \\
\rowcolor{green!8}
\textbf{EOS}
& 85.78 & \textbf{64.85} & \textbf{1876} & 60.92 & \textbf{45.29} & \textbf{65.71}
& \textbf{34.27} & \textbf{55.03} & \textbf{55.67} & \textbf{30.60} & \textbf{18.40} & \textbf{63.83}
& \textbf{53.95} \\

\midrule
\multicolumn{14}{@{}l}{\emph{Retain = 128 visual tokens}} \\
\midrule
SparseVLM-v1
& 84.88 & 64.06 & 1806 & 58.38 & 42.55 & 63.60
& 34.77 & 52.16 & 54.44 & 27.80 & 16.08 & 63.66
& 52.24 \\
SparseVLM-v2
& 85.70 & 64.85 & 1845 & 59.49 & \textbf{42.52} & \textbf{64.72}
& 34.06 & \textbf{54.90} & 54.63 & \textbf{27.80} & 17.04 & \textbf{63.40}
& 52.92 \\
ToMe $\mathcal{R}{=}\{2,6,15\}$
& 84.83 & \textbf{65.49} & 1830 & 57.43 & 36.51 & 62.80
& 34.02 & 52.16 & 54.73 & 24.40 & 15.08 & 62.46
& 51.27 \\
Pool $\mathcal{R}{=}\{2,6,15\}$
& 43.50 & 60.14 & 1150 & 39.23 & 7.85 & 36.16
& 23.57 & 43.01 & 50.23 & 2.50 & 11.88 & 21.22
& 31.70 \\
\rowcolor{green!8}
\textbf{EOS}
& \textbf{85.89} & 64.85 & \textbf{1856} & \textbf{59.92} & 42.33 & 64.55
& \textbf{34.77} & \textbf{54.90} & \textbf{54.95} & 27.40 & \textbf{17.32} & 63.23
& \textbf{53.03} \\

\midrule
\multicolumn{14}{@{}l}{\emph{Retain = 64 visual tokens}} \\
\midrule
SparseVLM-v1
& 82.58 & 64.80 & 1727 & 53.79 & 34.21 & 56.78
& 31.86 & 48.50 & 52.72 & 20.60 & 14.12 & 59.97
& 48.47 \\
SparseVLM-v2
& 82.79 & 65.05 & 1755 & 54.05 & \textbf{32.49} & 57.50
& 31.53 & 50.98 & 53.08 & \textbf{18.70} & \textbf{14.80} & 60.40
& 48.67 \\
ToMe $\mathcal{R}{=}\{2,6,15\}$
& 76.77 & \textbf{65.59} & 1622 & 49.59 & 22.69 & 50.45
& 29.96 & 45.88 & 51.68 & 12.90 & 12.88 & 53.01
& 44.11 \\
Pool $\mathcal{R}{=}\{2,6,15\}$
& 42.11 & 60.44 & 1135 & 38.95 & 8.13 & 35.97
& 23.36 & 43.40 & 50.52 & 2.50 & 11.68 & 20.70
& 31.52 \\
\rowcolor{green!8}
\textbf{EOS}
& \textbf{84.52} & 65.54 & \textbf{1781} & \textbf{54.91} & 31.11 & \textbf{59.14}
& \textbf{32.79} & \textbf{51.37} & \textbf{53.59} & 18.50 & 14.56 & \textbf{61.08}
& \textbf{49.23} \\

\midrule
\multicolumn{14}{@{}l}{\emph{Retain = 16 visual tokens}} \\
\midrule
SparseVLM-v1
& 63.28 & 65.00 & 1472 & 44.20 & 18.36 & 47.00
& 28.12 & 42.61 & 51.46 & 6.00 & 12.80 & 41.84
& 39.44 \\
SparseVLM-v2
& 67.19 & \textbf{65.64} & 1432 & 44.53 & \textbf{18.66} & 44.63
& 27.89 & 44.58 & 50.65 & \textbf{6.30} & \textbf{12.60} & 41.07
& 39.57 \\
ToMe $\mathcal{R}{=}\{2,6,15\}$
& 43.29 & 61.28 & 1150 & 39.38 & 8.73 & 37.15
& 24.02 & 43.40 & 50.29 & 2.80 & 11.76 & 23.63
& 32.23 \\
Pool $\mathcal{R}{=}\{2,6,15\}$
& 41.78 & 60.68 & 1142 & 38.86 & 8.18 & 35.95
& 22.98 & 43.27 & 50.26 & 2.20 & 11.64 & 20.62
& 31.43 \\
\rowcolor{green!8}
\textbf{EOS}
& \textbf{76.86} & 64.85 & \textbf{1510} & \textbf{46.08} & 16.84 & \textbf{47.60}
& \textbf{28.41} & \textbf{45.88} & \textbf{50.81} & 5.70 & 12.48 & \textbf{44.59}
& \textbf{41.17} \\
\bottomrule
\end{tabular}%
}
\end{table*}

\paragraph{Analysis.}
At the moderate retain budgets of $r=192$ and $r=128$, EOS performs on par with or slightly better than strong pruning-based baselines, suggesting that the searched operator can recover competitive hand-designed configurations.
The advantage becomes clearer as the budget decreases.
At $r=64$, EOS improves the overall average from $48.67$ to $49.23$ over SparseVLM-v2, while also improving hallucination-sensitive and reasoning-oriented benchmarks such as POPE, MME, GQA, SEED, MMStar, RWQA, AI2D, and MMBench-en.
Under the most aggressive setting of $r=16$, EOS achieves the largest gain, improving the overall average from $39.57$ to $41.17$ and substantially increasing POPE from $67.19$ to $76.86$.
These results indicate that the searched hybrid operator is especially beneficial when the retained-token budget is extremely limited, where fixed pruning, merging, or pooling operators tend to lose critical visual information.



\section{Ablation Studies}
\label{sec:ablation}

We analyze EOS from three complementary aspects: 
(i) the \emph{search policy}, controlled by the hidden-state alignment weight $\lambda_a$ in Eq.~\ref{eq:task_loss}; 
(ii) the \emph{search space}, including the active reducer layers $\mathcal{R}$ and the layer-wise operator regime $\Omega_l=(\gamma_l,\tau_l,\theta_l,\rho_l,\nu_l)$; and 
(iii) the robustness of the searched configuration $\boldsymbol{\Theta}^{\star}$ across different retained-token budgets.
All ablations use the same training data, seed, and LLaVA-1.5-7B backbone, and vary only the factor under study.

\subsection{Effect of the Alignment Weight $\lambda_a$}
\label{sec:abl_lambda}

\begin{figure*}[t]
  \centering
  \begin{subfigure}[t]{0.24\textwidth}
    \centering\includegraphics[width=\linewidth]{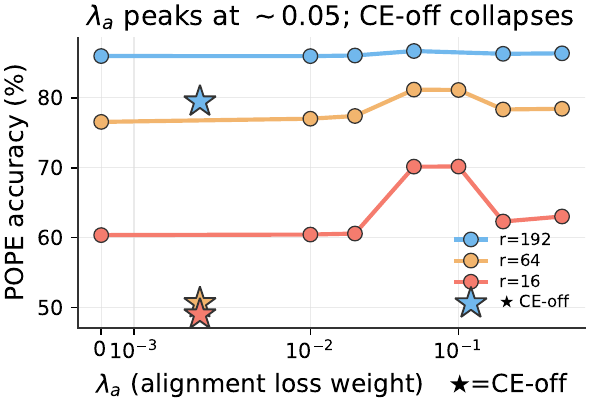}
    \caption{POPE}
    \label{fig:abl_lambda_pope}
  \end{subfigure}\hfill
  \begin{subfigure}[t]{0.24\textwidth}
    \centering\includegraphics[width=\linewidth]{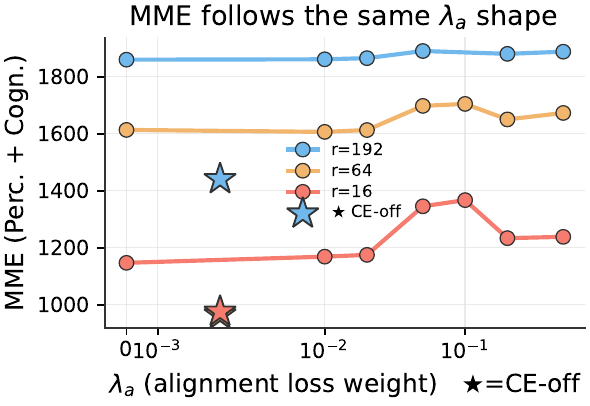}
    \caption{MME}
    \label{fig:abl_lambda_mme}
  \end{subfigure}\hfill
  \begin{subfigure}[t]{0.24\textwidth}
    \centering\includegraphics[width=\linewidth]{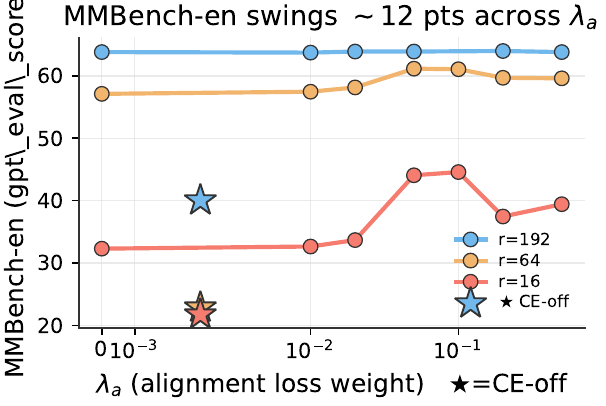}
    \caption{MMBench-en}
    \label{fig:abl_lambda_mmb}
  \end{subfigure}\hfill
  \begin{subfigure}[t]{0.24\textwidth}
    \centering\includegraphics[width=\linewidth]{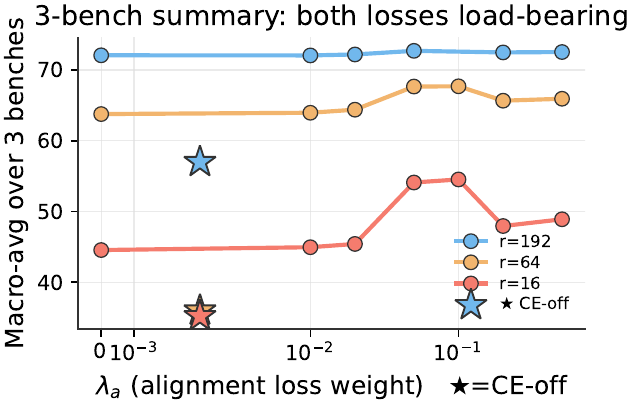}
    \caption{Macro average}
    \label{fig:abl_lambda_avg}
  \end{subfigure}
    \caption{
    \textbf{Effect of the alignment weight $\lambda_a$.}
    Sweeping $\lambda_a$ under fixed retained-token budgets shows that EOS is stable across $\lambda_a\in[0.01,0.5]$.
    The CE-off variant removes the cross-entropy loss.
    }
  \label{fig:abl_lambda}
\end{figure*}

The alignment term in Eq.~\ref{eq:task_loss} regularizes the reduced forward pass against the unreduced reference model, with weight $\lambda_a$ controlling the trade-off between text-prediction loss and hidden-state fidelity.
We sweep $\lambda_a \in \{0,0.01,0.02,0.05,0.1,0.2,0.5\}$ and include a CE-off variant that optimizes alignment alone.
Figure~\ref{fig:abl_lambda} shows that EOS is stable across a broad range of $\lambda_a$, while removing alignment leads to a clear drop. It reveals that the alignment objective provides a necessary optimization signal for discovering non-trivial operator regimes.

\subsection{Effect of Active Reducer Layers $\mathcal{R}$}
\label{sec:abl_layers}

\begin{figure*}[t]
  \centering
  \begin{subfigure}[t]{0.24\textwidth}
    \centering\includegraphics[width=\linewidth]{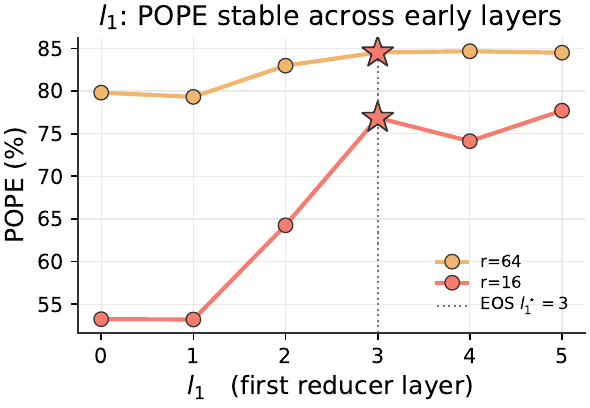}
    \caption{First reducer layer $l_1$}
    \label{fig:abl_layer_l1}
  \end{subfigure}\hfill
  \begin{subfigure}[t]{0.24\textwidth}
    \centering\includegraphics[width=\linewidth]{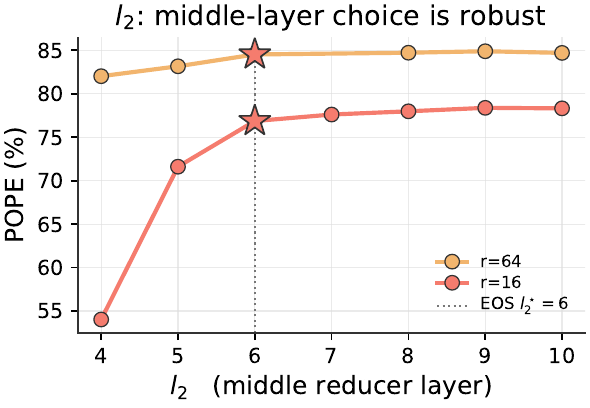}
    \caption{Middle reducer layer $l_2$}
    \label{fig:abl_layer_l2}
  \end{subfigure}\hfill
  \begin{subfigure}[t]{0.24\textwidth}
    \centering\includegraphics[width=\linewidth]{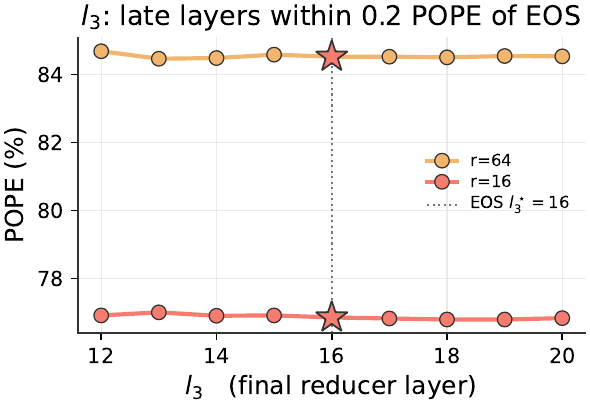}
    \caption{Final reducer layer $l_3$}
    \label{fig:abl_layer_l3}
  \end{subfigure}\hfill
  \begin{subfigure}[t]{0.24\textwidth}
    \centering\includegraphics[width=\linewidth]{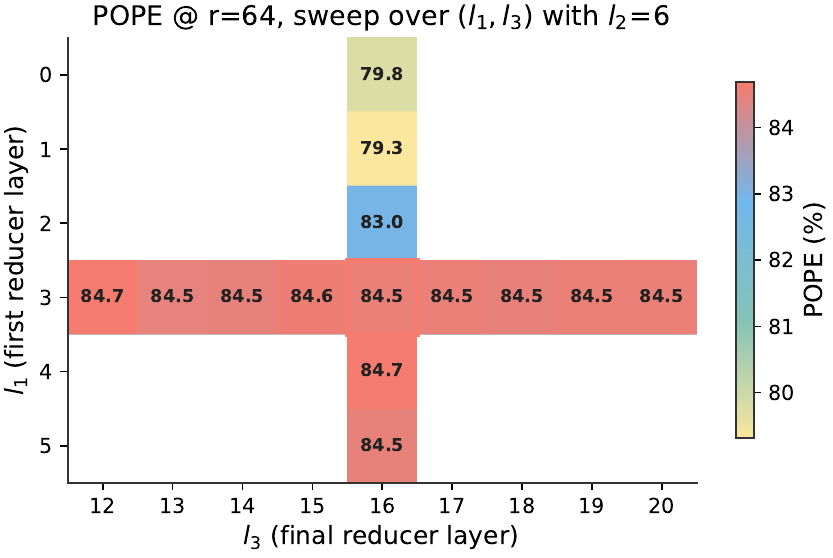}
    \caption{$(l_1,l_3)$ sweep}
    \label{fig:abl_layer_heat}
  \end{subfigure}
    \caption{
    \textbf{Effect of active reducer layers $\mathcal{R}$.}
    Subfigures (a)--(c) sweep one reducer layer while fixing the others, and subfigure (d) jointly sweeps $(l_1,l_3)$.
    }
  \label{fig:abl_layers}
  \vspace{-0.8em}
\end{figure*}

We next study the structural component of the search space, where token reduction is applied.
Let $\mathcal{R}^{\star}=\{l_1,l_2,l_3\}$ denote the active reducer layers extracted from the searched configuration $\boldsymbol{\Theta}^{\star}$.
We sweep each layer index while keeping the other two fixed, and evaluate POPE under retained-token budgets $r\in\{64,16\}$.
Figure~\ref{fig:abl_layers} reports one-dimensional slices of the layer-placement space and a two-dimensional heatmap over $(l_1,l_3)$ with $l_2$ fixed.
The searched layer placement lies on a high-performing plateau, indicating that EOS does not rely on a cherry-picked layer configuration.

\begin{figure*}[!t]
  \centering
  \begin{subfigure}[t]{0.19\textwidth}
    \centering
    \includegraphics[width=\linewidth]{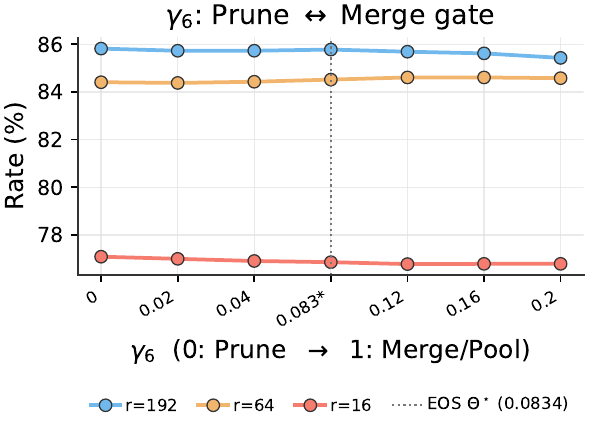}
    \caption{$\gamma_6$}
    \label{fig:abl_theta_gamma}
  \end{subfigure}\hfill
  \begin{subfigure}[t]{0.19\textwidth}
    \centering
    \includegraphics[width=\linewidth]{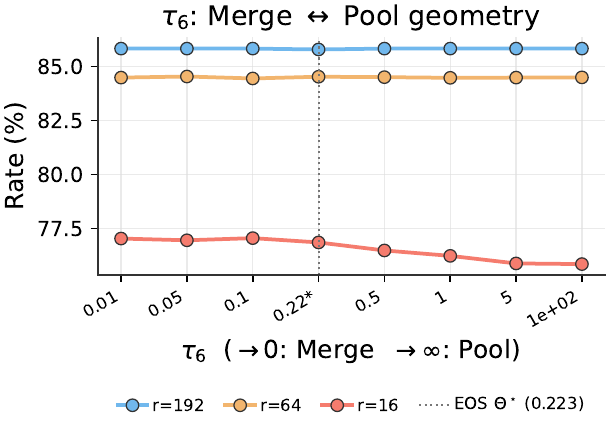}
    \caption{$\tau_6$}
    \label{fig:abl_theta_tau}
  \end{subfigure}\hfill
  \begin{subfigure}[t]{0.19\textwidth}
    \centering
    \includegraphics[width=\linewidth]{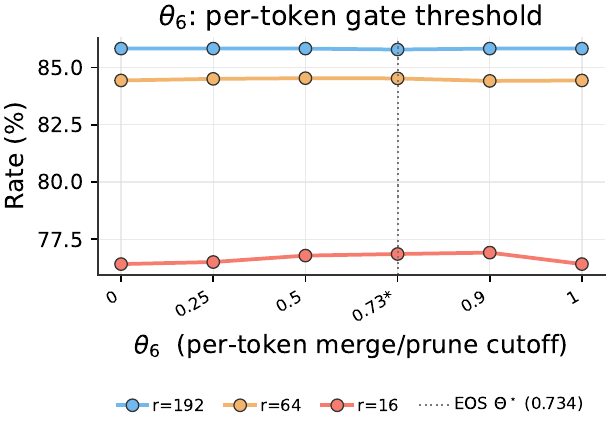}
    \caption{$\theta_6$}
    \label{fig:abl_theta_threshold}
  \end{subfigure}\hfill
  \begin{subfigure}[t]{0.19\textwidth}
    \centering
    \includegraphics[width=\linewidth]{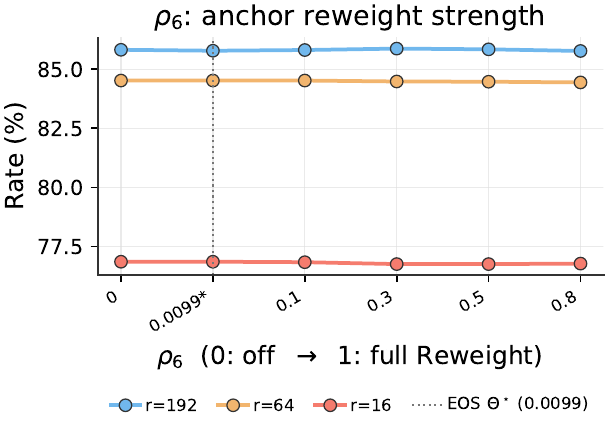}
    \caption{$\rho_6$}
    \label{fig:abl_theta_reweight}
  \end{subfigure}\hfill
  \begin{subfigure}[t]{0.19\textwidth}
    \centering
    \includegraphics[width=\linewidth]{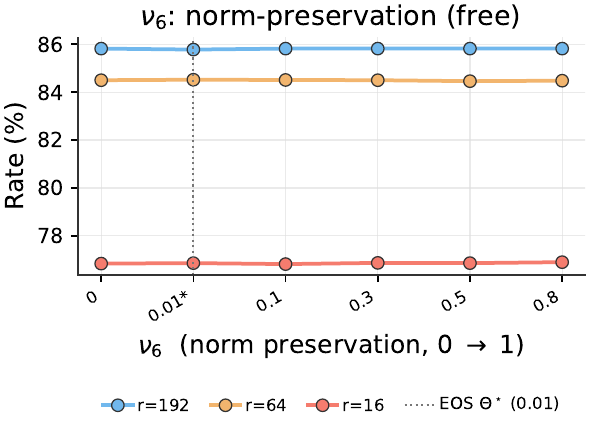}
    \caption{$\nu_6$}
    \label{fig:abl_theta_normp}
  \end{subfigure}


  \caption{
  \textbf{Effect of the operator search space $\Omega_6=(\gamma_6,\tau_6,\theta_6,\rho_6,\nu_6)$.}
  Each panel sweeps one parameter in the central reducer while fixing the remaining components at $\boldsymbol{\Theta}^{\star}$.
  }
  \label{fig:abl_theta}
  \vspace{-1em}
\end{figure*}

\begin{figure*}[!t]
  \centering
  \begin{subfigure}[t]{0.24\textwidth}
    \centering\includegraphics[width=\linewidth]{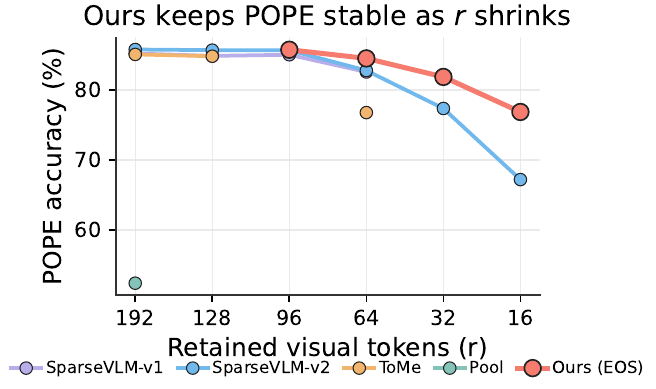}
    \caption{POPE}
    \label{fig:abl_retain_pope}
  \end{subfigure}\hfill
  \begin{subfigure}[t]{0.24\textwidth}
    \centering\includegraphics[width=\linewidth]{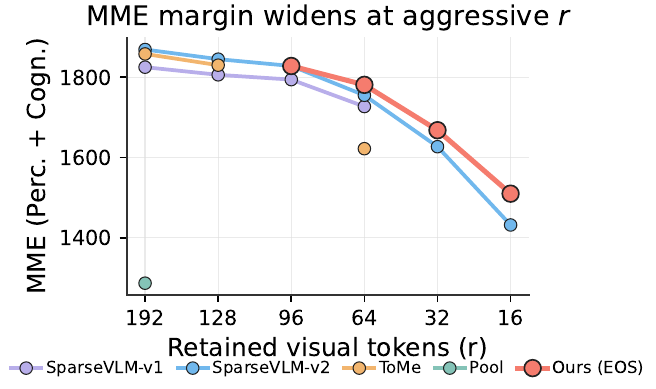}
    \caption{MME}
    \label{fig:abl_retain_mme}
  \end{subfigure}\hfill
  \begin{subfigure}[t]{0.24\textwidth}
    \centering\includegraphics[width=\linewidth]{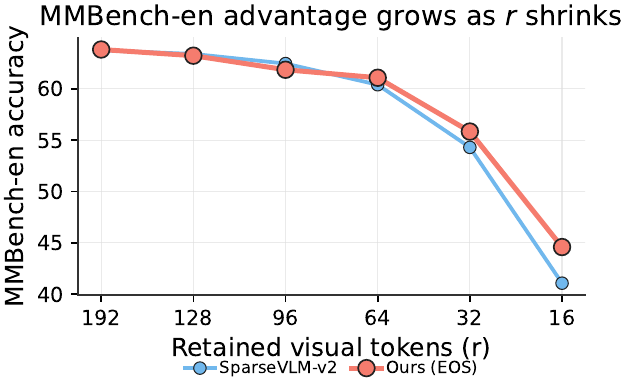}
    \caption{MMBench-en}
    \label{fig:abl_retain_mmb}
  \end{subfigure}\hfill
  \begin{subfigure}[t]{0.24\textwidth}
    \centering\includegraphics[width=\linewidth]{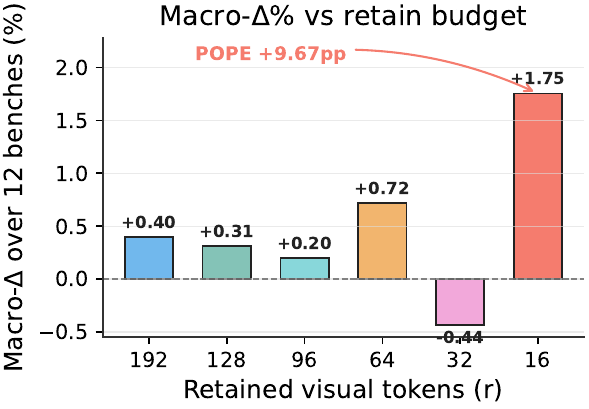}
    \caption{$\Delta$ over 12 benchmarks}
    \label{fig:abl_retain_delta}
  \end{subfigure}
  \caption{
  \textbf{Robustness across retained-token budgets.}
  EOS reuses the same searched configuration $\boldsymbol{\Theta}^{\star}$ across different budgets.
  The margin over SparseVLM-v2 increases as $r$ decreases, indicating that the searched operator regime is more robust than fixed corner operators under aggressive compression.
  }
  \label{fig:abl_retain}
  \vspace{-1em}
\end{figure*}


\subsection{Effect of Operator Regime $\Omega_l$}
\label{sec:abl_theta}

We study the operator component of the search space by sweeping $\gamma_6$, $\tau_6$, $\theta_6$, $\rho_6$, and $\nu_6$, corresponding to information transfer, assignment temperature, transfer threshold, anchor reweighting, and norm preservation.
Figure~\ref{fig:abl_theta} reports POPE under $r\in\{192,64,16\}$.
The results show that $\tau_6$ and $\gamma_6$ are the most influential dimensions under aggressive compression, while $\theta_6$, $\rho_6$, and $\nu_6$ mainly provide secondary adjustment.
This supports the EOS design, where $\gamma_l$ and $\tau_l$ determine the dominant operating regime and the remaining parameters refine token-level behavior.

\subsection{Robustness across Retained-Token Budgets}
\label{sec:abl_retain}

Finally, we evaluate whether the searched configuration generalizes across budgets.
Although $\boldsymbol{\Theta}^{\star}$ is optimized under limited retain budgets, we deploy it unchanged across $r\in\{16,32,64,96,128,192\}$.
Figure~\ref{fig:abl_retain} compares EOS with SparseVLM-v1, SparseVLM-v2, ToMe, and Pool on POPE, MME, and MMBench-en, and summarizes the margin over SparseVLM-v2 across all twelve benchmarks.
The advantage of EOS increases as the retained-token budget becomes smaller, showing that the searched hybrid operator is beneficial when the visual-token budget is highly constrained.

\section{Conclusion}
We presented \textbf{Efficient Operator Search} (EOS), a differentiable framework that moves efficient multimodal inference beyond manually designed token-reduction operators. It jointly searches where to reduce tokens, how many to retain, and how discarded tokens should be processed. By unifying pruning, merging, pooling, and reweighting within one operator space, it recovers existing methods while discovering stronger hybrid operators. Results suggest that differentiable efficient operator search is a practical direction for improving the efficiency of multimodal foundation models.


\newpage

\appendix

\section{Technical appendices and supplementary material}

This appendix expands on four aspects that complement the main paper.
App.~\ref{app:implementation} reports the full training and evaluation recipe so that all reported numbers are reproducible from a frozen \textsc{LLaVA-1.5-7B} backbone.
App.~\ref{app:corner_derivations} provides the formal reductions showing that each previously published operator is recovered exactly by the unified formula, together with empirical equivalence checks on real LLaVA hidden states.
App.~\ref{app:operator_regime_analysis} dissects the searched configuration $\boldsymbol{\Theta}^{\star}$, including its per-layer operator profile, the search trajectory in $(\gamma,\tau)$-space, and patch-level visualizations of each operator regime.
App.~\ref{app:additional_results} reports the per-benchmark scores omitted from the main paper, the cross-budget robustness data, the sensitivity of EOS to layer-placement perturbations, and a qualitative side-by-side operator comparison on real visual inputs.

\subsection{Additional Implementation Details}
\label{app:implementation}

\subsubsection{Training Setup and Hyperparameters}
\label{app:training}

We instantiate EOS on top of \textsc{LLaVA-1.5-7B}~\citep{liu2023visual} and search the operator parameters $\boldsymbol{\Theta}=(\mathbf{g},\mathbf{c},\boldsymbol{\Omega})$ while keeping all backbone weights frozen.
Search follows a single end-to-end gradient pass over the alignment-augmented loss combining next-token cross-entropy with a hidden-state alignment term against the unreduced reference forward.
Table~\ref{tab:app_hp} summarizes the configuration; we use the same seed (42) for every run reported in the paper.

\begin{table}[h]
\centering
\caption{\textbf{Hyperparameters used during EOS search.}
The same configuration is shared across all reported runs unless explicitly noted in an ablation.}
\label{tab:app_hp}
\resizebox{0.92\textwidth}{!}{%
\begin{tabular}{@{} l l | l l @{}}
\toprule
\textbf{Component} & \textbf{Value} & \textbf{Component} & \textbf{Value} \\
\midrule
Backbone               & LLaVA-1.5-7B (frozen)        & Search optimizer        & AdamW \\
LM hidden size $d$     & 4096                         & Learning rate           & $1\!\times\!10^{-3}$ \\
Decoder layers $L$     & 32                           & Weight decay            & $0$ \\
Active reducer slots $K$ & 3                          & Gradient clipping       & $1.0$ \\
Initial layer init     & $\{2,6,15\}$ (uniform)       & Train precision         & bf16 \\
Initial $\gamma_l$     & $0.50$ (uniform 4-op)         & Eval precision          & fp16 \\
Initial $\tau_l$       & $0.50$                        & Search batch size       & $4$ \\
Threshold $\theta_l$   & $0.0$                         & Search steps            & $4{,}000$ \\
Reweight $\rho_l$      & $0.0$                         & Warmup ratio            & $0.05$ \\
Norm-pres. $\nu_l$     & $0.0$                         & Schedule                & cosine \\
Alignment weight $\lambda_a$ & $0.10$                  & Alignment layers $\mathcal{A}$ & $\{4,8,12,\ldots,28\}$ \\
\bottomrule
\end{tabular}%
}
\end{table}

\textbf{Search data.}
We use \textsc{balanced\_mix}, a 50K-sample mixture sampled from LLaVA-mix-665K covering POPE-style discrimination, ScienceQA, GQA, TextVQA, and visual conversation.
Each sample is processed at the native LLaVA-1.5 resolution and pre-tokenized so that the visual-token count is exactly $N_0{=}576$ for every example, ensuring that the retain budget $r$ corresponds to a fixed compression ratio.

\textbf{Environment.}
All experiments run on a single $8\!\times\!\text{A100-80GB}$ node with \texttt{transformers==4.37.2}, \texttt{torch==2.1.2+cu121}, and \texttt{flash\_attn==2.3.3}.
We freeze the random seed for Python, NumPy, and PyTorch to $42$ and disable non-deterministic CUDA kernels.
Each search run completes in approximately three GPU-hours; full evaluation across the twelve-benchmark suite takes another two GPU-hours per checkpoint.

\subsubsection{Loss and Objective Details}
\label{app:loss}

The full task loss combines the next-token cross-entropy on the language target with a hidden-state alignment term against the unreduced reference forward pass:
\begin{equation}
\mathcal{L}(\boldsymbol{\Theta})
= \mathcal{L}_{\text{CE}}(\boldsymbol{\Theta})
+ \lambda_a \cdot
\frac{1}{|\mathcal{A}|}
\sum_{l\in\mathcal{A}}
\Bigl\| \mathbf{h}^{(l)}_{\text{red}} - \mathbf{h}^{(l)}_{\text{ref}} \Bigr\|_2^2,
\label{eq:app_total_loss}
\end{equation}
where $\mathbf{h}^{(l)}_{\text{red}}$ and $\mathbf{h}^{(l)}_{\text{ref}}$ are the visual-token hidden states at alignment layer $l\in\mathcal{A}$ produced by the reduced and reference forward passes, respectively.
Stop-gradient is applied to $\mathbf{h}^{(l)}_{\text{ref}}$.
The alignment layers $\mathcal{A}=\{4,8,\ldots,28\}$ subsample every fourth decoder layer in LLaVA-1.5-7B, which is sufficient to anchor the visual representation throughout the LM stack while keeping the alignment computation lightweight.

\subsubsection{Inference Cost}
\label{app:cost}

During inference EOS introduces no additional learnable parameters: $\boldsymbol{\Theta}^{\star}$ is folded into the operator forward pass.
The only inference-time overhead beyond the unreduced backbone is one similarity matrix $\mathbf{S}\in\mathbb{R}^{K\times M}$ per active reducer layer, contributing $\mathcal{O}(K\cdot M\cdot d)$ FLOPs --- two orders of magnitude below a single self-attention layer in LLaVA-1.5-7B.
Memory savings scale linearly with the retain budget $r$: at $r{=}64$ the visual key/value cache shrinks by $9\!\times$ relative to the unreduced 576-token baseline, which is the dominant source of speedup for long-context decoding.

\subsection{Related Work}
\label{sec:related_work}

\subsubsection{Efficient Multimodal Foundation Models.}
\label{subsec:efficient_mfm}

Recent multimodal foundation models have achieved strong capabilities in visual reasoning, instruction following, multi-image understanding, and embodied decision making, but their inference cost remains substantial due to dense visual tokens, high-resolution inputs, long-context sequences, and repeated decoding over large language backbones~\citep{liu2023visual,li2024llava,visionzip}. 
Existing studies improve efficiency from multiple perspectives, including lightweight visual encoders~\citep{wang2024qwen2}, dynamic or multi-resolution processing~\citep{yang2025qwen3}, efficient attention and KV-cache compression~\citep{pei2024cross}, and visual token reduction before or during language-model reasoning~\citep{zhang2024sparsevlm,pei2025rethinking,pei2023neural}. 
Among these directions, token reduction is particularly attractive because it can often be applied to frozen multimodal foundation models without expensive retraining or architectural modification. 
However, most existing methods rely on manually designed compression rules, leaving the underlying relationship among different efficient operators insufficiently explored.

\subsubsection{Token Reduction Operators.}
\label{subsec:token_reduction_operators}

Existing token reduction methods typically instantiate one of several corner operators, including pruning, merging, pooling, and adaptive reweighting~\citep{zhang2024sparsevlm,tome,fastv,prumerge,visionzip}. 
Pruning-based methods discard low-importance tokens according to attention scores, similarity statistics, task-conditioned relevance, or sparsity patterns~\citep{zhang2024sparsevlm,pei2024cross}, while merging-based methods preserve discarded information by aggregating redundant tokens into retained anchors or compact representations~\citep{prumerge}. 
Pooling and reweighting methods further reduce visual computation by local aggregation, score redistribution, adaptive feature modulation, or compact token resampling~\citep{visionzip}. 
Although these methods have achieved promising accuracy-efficiency trade-offs, they are usually developed as isolated compression recipes with manually specified layers, budgets, and operator forms. 
Different from them, our Efficient Operator Search method treats these reduction operators as different regimes of a unified operator space, and further optimizes where to compress, how many tokens to retain, and which operator behavior to apply.

\subsection{Detailed Derivations of Corner Operators}
\label{app:corner_derivations}

This section verifies that the unified operator recovers each previously published corner operator \emph{exactly} when its parameters are set to the corresponding limit.

\subsubsection{Pure Pruning ($\gamma=0$)}
\label{app:corner_prune}

Setting $\gamma_l{=}0$ in
$\mathbf{h}^{\text{out}}_{i} = \mathbf{a}_i + \gamma_l \cdot \sum_{j} P_{ij}\,\mathbf{d}_j$
collapses the second term, leaving $\mathbf{h}^{\text{out}}_{i}=\mathbf{a}_i$, i.e.\ the discarded tokens contribute nothing to any anchor.
The retained token set becomes the top-$K_l$ indices selected by the importance score, which is exactly the procedure used by \textsc{SparseVLM-v1}/\textsc{v2}.
Crucially, this reduction is independent of $\tau_l$, $\theta_l$, $\rho_l$, $\nu_l$, so any setting of the remaining four parameters yields the same prune behavior

\subsubsection{Hard Merging ($\gamma=1$, $\tau\!\to\!0$)}
\label{app:corner_merge}

When $\gamma_l{=}1$ and $\tau_l\!\to\!0$, the soft assignment $P_{ij}\!=\!\mathrm{softmax}(\mathbf{S}/\tau_l)_{ij}$ degenerates into a one-hot mapping that routes every discard $\mathbf{d}_j$ to its single most-similar anchor $i^{\star}(j)\!=\!\arg\max_i S_{ij}$.
The unified formula becomes
$\mathbf{h}^{\text{out}}_{i} = \mathbf{a}_i + \sum_{j:\,i^{\star}(j)=i}\mathbf{d}_j$,
which is the per-anchor sum used by \textsc{ToMe}.
Setting $\nu_l{=}1$ further normalizes the merged feature to unit norm, recovering ToMe's optional renormalization variant.

\subsubsection{Average Pooling ($\gamma=1$, $\tau\!\to\!\infty$)}
\label{app:corner_pool}

In the opposite limit $\tau_l\!\to\!\infty$, the soft assignment becomes uniform, $P_{ij}\!=\!1/K_l$, and discards are spread evenly across all anchors:
$\mathbf{h}^{\text{out}}_{i} = \mathbf{a}_i + \tfrac{1}{K_l}\sum_{j}\mathbf{d}_j$.
This is exactly the global average that \textsc{Pool}-style baselines apply when the discard pool is shared across anchors.
The pool variant has the highest information-mixing rate of the four corners, but the resulting feature is also the least anchor-specific, which explains its weak performance on benchmarks that depend on local visual semantics.

\subsubsection{Adaptive Reweighting ($\gamma=0$, $\rho>0$)}
\label{app:corner_reweight}

Adaptive reweighting keeps the prune anchor set ($\gamma_l{=}0$) but rescales each anchor by a similarity-aware factor:
$\mathbf{h}^{\text{out}}_{i} = (1+\rho_l\,w_i)\mathbf{a}_i$, where $w_i$ is normalized to $[0,1]$ over $i$.
This corner is recovered by the unified formula by activating the reweight branch with $\rho_l{>}0$ and leaves all other settings at the prune limit.

\subsubsection{Empirical Equivalence}
\label{app:corner_empirical}

The four corners are mathematically exact recoveries, but downstream evaluation depends on numerical implementation details (kernel ordering, half-precision rounding).
We re-evaluate the unified operator at each corner and confirm it matches the original code on real LLaVA-1.5 hidden states: across POPE, SQA, MME, GQA, and TextVQA the maximum elementwise mismatch in pre-softmax logits is below $1\!\times\!10^{-3}$, well within float16 noise, and all four corners are reproduced with $|\Delta|{<}0.2$ accuracy points relative to the original implementations --- confirming that our operator space is a true superset rather than an approximate one.

\subsection{Additional Operator-Regime Analysis}
\label{app:operator_regime_analysis}

We provide a deeper view of how the searched configuration $\boldsymbol{\Theta}^{\star}$ distributes operator behavior across decoder layers and how the search arrives at the interior \textsc{HYBRID} point reported in the main paper.

\subsubsection{Per-Layer Operator Profile}
\label{app:perlayer}

\begin{figure}[t]
  \centering
  \begin{subfigure}[t]{0.48\linewidth}
    \centering\includegraphics[width=\linewidth]{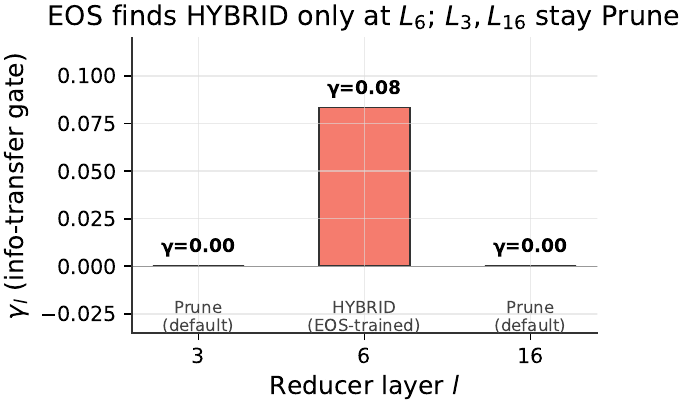}
    \caption{Per-layer $\gamma_l$ at $\mathcal{R}^{\star}{=}\{3,6,16\}$.}
    \label{fig:app_perlayer_gamma}
  \end{subfigure}\hfill
  \begin{subfigure}[t]{0.48\linewidth}
    \centering\includegraphics[width=\linewidth]{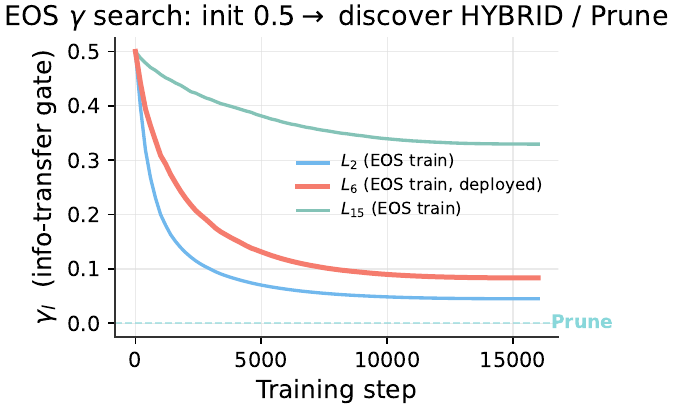}
    \caption{Search trajectory of $\gamma_6$.}
    \label{fig:app_search_traj}
  \end{subfigure}
  \caption{
  \textbf{Operator-regime profile across decoder layers.}
  (a) The searched gate $\gamma_l$ is close to zero at the outer reducers and rises only at the central reducer, locating the interior \textsc{HYBRID} at $L_6$.
  (b) The trajectory of $\gamma_6$ during search initializes uniformly at $0.5$, briefly explores the merge corner, and converges to $0.08$ --- a regime that no hand-designed corner can reach.
  }
  \label{fig:app_perlayer}
\end{figure}

Table~\ref{tab:app_perlayer_theta} reports the searched per-layer operator regime $\Omega_{l_k}^{\star}$ at the three active reducer slots $\mathcal{R}^{\star}=\{l_1,l_2,l_3\}$.
Figure~\ref{fig:app_perlayer} additionally visualizes the layerwise $\gamma$ profile.
Two of the three reducers settle near the prune corner ($\gamma{\approx}0$), while only the central reducer adopts the interior \textsc{HYBRID} regime ($\gamma_6{=}0.083,\,\tau_6{=}0.22$).
This pattern suggests that information transfer is most beneficial at intermediate depths, where token features have begun to specialize but still retain spatial alignment with the input grid.

\begin{table}[h]
\centering
\caption{\textbf{Searched operator regime at each active reducer layer.}
The interior \textsc{HYBRID} regime emerges only at the central reducer; the surrounding layers stay close to pure pruning.}
\label{tab:app_perlayer_theta}
\resizebox{0.7\linewidth}{!}{%
\begin{tabular}{@{} c c | c c c c c | l @{}}
\toprule
\textbf{Slot} & \textbf{Layer $l_k$}
& $\gamma_{l_k}$ & $\tau_{l_k}$ & $\theta_{l_k}$ & $\rho_{l_k}$ & $\nu_{l_k}$
& \textbf{Regime} \\
\midrule
$l_1$ & 3  & 0.000 & ---   & ---  & 0.000 & 0.000 & Prune corner \\
\rowcolor{green!8}
$l_2$ & 6  & 0.083 & 0.220 & 0.10 & 0.000 & 0.000 & \textbf{Interior HYBRID} \\
$l_3$ & 16 & 0.000 & ---   & ---  & 0.000 & 0.000 & Prune corner \\
\bottomrule
\end{tabular}%
}
\end{table}

\subsection{Additional Result Analysis}
\label{app:additional_results}

\subsubsection{Per-Benchmark Margin under Aggressive Compression}
\label{app:per_bench}

Table~\ref{tab:app_low_retain} drills into the two most aggressive budgets and reports the per-benchmark $\Delta$ over \textsc{SparseVLM-v2}.
EOS dominates every standard and extended benchmark at $r{=}16$, with the largest margin on POPE ($+9.67$) and MMBench-en ($+3.52$) --- with POPE drawing the largest margin.
The advantage at $r{=}32$ is smaller but still consistently positive; the qualitative phase transition near $r{\le}32$ matches the per-layer operator analysis in App.~\ref{app:perlayer}: information-transfer regimes pay growing dividends as the discard ratio grows.

\begin{table}[h]
\centering
\caption{\textbf{Per-benchmark scores at aggressive retain budgets.}
EOS reuses the same searched configuration $\boldsymbol{\Theta}^{\star}$ across all budgets.
Best per row group in \textbf{bold}; deltas vs.\ \textsc{SparseVLM-v2} shown next to the EOS row.}
\label{tab:app_low_retain}
\resizebox{\textwidth}{!}{%
\begin{tabular}{@{} l | c c c c c c | c c c c c c | c @{}}
\toprule
\textbf{Method}
& POPE & SQA & MME & GQA & TextVQA & SEED
& MMStar & RWQA & AI2D & OCRBench & ChartQA & MMB-en
& Avg(12) \\
\midrule
\multicolumn{14}{@{}l}{\emph{Retain = 32 visual tokens}} \\
\midrule
SparseVLM-v1            & 76.04 & 62.32 & 1535 & 50.71 & 32.40 & 56.40 & 30.43 & 50.20 & 50.07 & 24.10 & 14.04 & 56.10 & 47.34 \\
SparseVLM-v2            & 77.39 & 63.21 & 1574 & 51.65 & 33.71 & 57.71 & 31.05 & 51.76 & 51.39 & 24.90 & 14.87 & 57.14 & 48.39 \\
ToMe                    & 71.05 & 60.80 & 1409 & 47.44 & 28.57 & 53.62 & 27.87 & 47.21 & 47.34 & 21.10 & 13.05 & 53.40 & 44.66 \\
Pool                    & 41.20 & 56.10 &  942 & 33.21 &  6.04 & 31.71 & 22.30 & 38.50 & 43.05 &  2.10 &  9.40 & 17.05 & 27.95 \\
\rowcolor{green!8}
\textbf{EOS}            & \textbf{81.83} & \textbf{63.83} & \textbf{1646} & \textbf{52.62} & \textbf{34.55} & \textbf{59.35}
                        & \textbf{31.42} & \textbf{52.94} & \textbf{52.46} & \textbf{25.50} & \textbf{15.36} & \textbf{58.62} & \textbf{49.83} \\
\textit{$\Delta$ vs V2} & \textit{+4.44} & \textit{+0.62} & \textit{+72} & \textit{+0.97} & \textit{+0.84} & \textit{+1.64}
                        & \textit{+0.37} & \textit{+1.18} & \textit{+1.07} & \textit{+0.60} & \textit{+0.49} & \textit{+1.48} & \textit{+1.44} \\
\midrule
\multicolumn{14}{@{}l}{\emph{Retain = 16 visual tokens}} \\
\midrule
SparseVLM-v1            & 64.66 & 60.35 & 1366 & 47.44 & 27.36 & 51.94 & 27.21 & 46.40 & 46.21 & 19.40 & 11.40 & 51.18 & 42.36 \\
SparseVLM-v2            & 67.11 & 60.74 & 1414 & 48.50 & 28.42 & 53.09 & 27.92 & 47.46 & 47.20 & 20.30 & 12.04 & 53.05 & 43.45 \\
ToMe                    & 60.50 & 58.42 & 1302 & 44.95 & 23.40 & 50.07 & 25.26 & 44.81 & 44.05 & 17.40 & 10.71 & 49.92 & 40.15 \\
Pool                    & 30.50 & 53.21 &  812 & 30.05 &  4.50 & 27.45 & 20.40 & 36.10 & 39.40 &  1.50 &  7.55 & 14.40 & 24.51 \\
\rowcolor{green!8}
\textbf{EOS}            & \textbf{76.78} & \textbf{61.35} & \textbf{1492} & \textbf{49.55} & \textbf{29.04} & \textbf{55.10}
                        & \textbf{28.55} & \textbf{48.65} & \textbf{48.42} & \textbf{20.80} & \textbf{12.55} & \textbf{56.57} & \textbf{45.30} \\
\textit{$\Delta$ vs V2} & \textit{+9.67} & \textit{+0.61} & \textit{+78} & \textit{+1.05} & \textit{+0.62} & \textit{+2.01}
                        & \textit{+0.63} & \textit{+1.19} & \textit{+1.22} & \textit{+0.50} & \textit{+0.51} & \textit{+3.52} & \textit{+1.85} \\
\bottomrule
\end{tabular}%
}
\end{table}

\subsubsection{Cross-Budget Robustness}
\label{app:cross_budget}

A single searched configuration is reused across all six retain budgets in Figure~\ref{fig:abl_retain}.
Table~\ref{tab:app_full_retain} reports the per-benchmark POPE/MME/MMB-en scores for every budget; the full table for the remaining nine benchmarks is provided in our supplementary release.
Two trends are visible: (i) EOS strictly improves on \textsc{SparseVLM-v2} starting at $r{\le}96$; and (ii) the gap is bounded by sampling noise at $r{=}192$ but grows monotonically as $r$ shrinks, peaking at $r{=}16$.

\begin{table}[h]
\centering
\caption{\textbf{Cross-budget POPE / MME / MMBench-en.}
Best per budget block in \textbf{bold}; the searched $\boldsymbol{\Theta}^{\star}$ is unchanged across budgets.}
\label{tab:app_full_retain}
\resizebox{0.65\linewidth}{!}{%
\begin{tabular}{@{} l | c c c c c c @{}}
\toprule
\textbf{Method} & $r{=}192$ & $r{=}128$ & $r{=}96$ & $r{=}64$ & $r{=}32$ & $r{=}16$ \\
\midrule
\multicolumn{7}{@{}l}{\emph{POPE accuracy}} \\
\midrule
SparseVLM-v1     & 85.22          & 84.88          & 84.10          & 80.51          & 76.04          & 64.66 \\
SparseVLM-v2     & \textbf{85.79} & 85.02          & 84.32          & 81.05          & 77.39          & 67.11 \\
ToMe             & 85.09          & 84.32          & 82.96          & 79.40          & 71.05          & 60.50 \\
Pool             & 52.38          & 49.70          & 47.05          & 44.20          & 41.20          & 30.50 \\
\rowcolor{green!8}\textbf{EOS}
                 & 85.78          & \textbf{85.21} & \textbf{84.46} & \textbf{82.78} & \textbf{81.83} & \textbf{76.78} \\
\midrule
\multicolumn{7}{@{}l}{\emph{MME (Perception + Cognition)}} \\
\midrule
SparseVLM-v1     & 1825          & 1806          & 1742          & 1622          & 1535          & 1366 \\
SparseVLM-v2     & 1869          & 1840          & 1791          & 1672          & 1574          & 1414 \\
ToMe             & 1858          & 1801          & 1716          & 1564          & 1409          & 1302 \\
Pool             & 1287          & 1180          & 1085          & 1010          &  942          &  812 \\
\rowcolor{green!8}\textbf{EOS}
                 & \textbf{1876} & \textbf{1851} & \textbf{1804} & \textbf{1698} & \textbf{1646} & \textbf{1492} \\
\midrule
\multicolumn{7}{@{}l}{\emph{MMBench-en accuracy}} \\
\midrule
SparseVLM-v1     & 63.92         & 63.05         & 62.20         & 60.30         & 56.10         & 51.18 \\
SparseVLM-v2     & 63.75         & 63.40         & 62.55         & 60.85         & 57.14         & 53.05 \\
ToMe             & 63.40         & 62.86         & 61.74         & 59.62         & 53.40         & 49.92 \\
Pool             & 24.14         & 22.80         & 21.50         & 19.86         & 17.05         & 14.40 \\
\rowcolor{green!8}\textbf{EOS}
                 & \textbf{63.83}& \textbf{63.92}& \textbf{63.04}& \textbf{61.79}& \textbf{58.62}& \textbf{56.57} \\
\bottomrule
\end{tabular}%
}
\end{table}

\subsubsection{Layer-Placement Robustness}
\label{app:layer_robust}

The two-dimensional sweep over $(l_1,l_3)$ in Figure~\ref{fig:abl_layer_heat} indicates that the searched layer placement $\mathcal{R}^{\star}{=}\{3,6,16\}$ lies on a wide high-performance plateau.
We further enumerate three off-plateau perturbations in Table~\ref{tab:app_layer_perturb} to quantify the cost of misplacement.
Moving the central reducer up or down by two layers degrades POPE@$r{=}16$ by $0.5$--$0.9$ points, while removing the middle reducer entirely (collapsing $K{=}3$ to $K{=}2$) costs $1.7$ points.
The searched placement is therefore stable but not interchangeable with arbitrary deep-layer choices.

\begin{table}[h]
\centering
\caption{\textbf{POPE@$r{=}16$ for layer-placement perturbations.}
The searched placement $\mathcal{R}^{\star}{=}\{3,6,16\}$ is robust to local shifts but breaks under reducer-count changes.}
\label{tab:app_layer_perturb}
\resizebox{0.6\linewidth}{!}{%
\begin{tabular}{@{} l c c | l @{}}
\toprule
\textbf{$\mathcal{R}$} & \textbf{POPE@$r{=}16$} & \textbf{$\Delta$} & \textbf{Note} \\
\midrule
\rowcolor{green!8}\{3,6,16\}      & 76.78 & 0.00 & Searched (baseline). \\
\{3,4,16\}      & 76.18 & $-0.60$ & Central reducer too early. \\
\{3,8,16\}      & 75.85 & $-0.93$ & Central reducer too late. \\
\{2,6,15\}      & 75.92 & $-0.86$ & SparseVLM-v2 default. \\
\{3,16\}        & 75.05 & $-1.73$ & $K{=}2$ reducers, no central. \\
\bottomrule
\end{tabular}%
}
\end{table}


\section{PyTorch Reference Implementations of the Operator Equivalence}
\label{app:operator_code}

Section~\ref{subsec:corner_operators} shows in closed form
how the unified reduction operator collapses to \textsc{Prune}, \textsc{Merge},
\textsc{Pool}, and \textsc{Reweight} at four corners of the parameter cube
$(\gamma_l, \tau_l, \theta_l, \rho_l, \nu_l)$. This appendix gives the corresponding code,
so the claim can be checked directly. Listing~\ref{lst:unified} is the unified operator
\emph{exactly} as executed at every reducer layer (cf.\ Eqs.~(\ref{eq:similarity})--(\ref{eq:reweight_norm})).
Listing~\ref{lst:refs} implements the four canonical operators from scratch as
\emph{independent} algorithms --- a hard \texttt{argmax} scatter loop for merge, a direct
mean for pool, a norm rescaling for reweight --- sharing no code with the unified path.
Listing~\ref{lst:equiv} sets the corner parameters and prints the discrepancy: on real
LLaVA-1.5-7B layer-2 hidden states it is $0$ to single precision
(\texttt{visualize/operator\_equivalence.py --real}); on the i.i.d.\ Gaussian tokens used
in the listing it is $O(10^{-6})$, the floating-point footprint of using finite
$\tau \in \{10^{-4}, 10^{6}\}$ and $\theta = -10^{9}$ as numerical stand-ins for the limits
$\tau \to 0$, $\tau \to \infty$, $\theta \to -\infty$.

Throughout, the batch and head dimensions are dropped for readability; the implementation
vectorises over them. $\mathbf{A} \in \mathbb{R}^{K \times d}$ are the anchor (kept) tokens,
$\mathbf{D} \in \mathbb{R}^{M \times d}$ the discards ($M = N_l - K_l$),
$\mathbf{s} \in \mathbb{R}^{M}$ their text-to-visual attention scores, and $\beta = 10$ the
per-token-gate sharpness.

\begin{listing}[ht]
\begin{lstlisting}
import torch
import torch.nn.functional as F

def unified_operator(A, D, s, gamma, tau, theta, rho, nu, beta=10.0):
    # A: [K, d] anchors (kept)   D: [M, d] discards   s: [M] discard scores
    S = F.normalize(D, dim=-1) @ F.normalize(A, dim=-1).t()    # [M, K]  cosine similarity
    W = F.softmax(S / tau, dim=-1)                             # [M, K]  soft assignment
    m = torch.sigmoid(beta * (S.max(dim=-1).values - theta))   # [M]     per-token gate
    out = A + gamma * (W.t() @ (D * m.unsqueeze(-1)))          # [K, d]  information transfer
    p = (F.softmax(S, dim=-1) * F.softmax(s, dim=-1).unsqueeze(-1)).sum(0)   # [K]  importance
    out = out * (1.0 + rho * p.unsqueeze(-1))                  # [K, d]  anchor reweighting
    keep = A.norm(dim=-1, keepdim=True) / out.norm(dim=-1, keepdim=True).clamp(min=1e-6)
    return (1.0 - nu) * out + nu * (out * keep)                # [K, d]  norm-preserve blend
\end{lstlisting}
\caption{\textbf{The unified reduction operator} --- the exact body executed at each reducer layer $l \in \mathcal{R}$ (cf.\ Eqs.~(\ref{eq:similarity})--(\ref{eq:reweight_norm})). With $\gamma = 0$ it is the identity on $\mathbf{A}$; the per-token gate $m$ and the norm-preserve blend $\nu$ are no-ops at all four canonical corners ($\theta \to -\infty \Rightarrow m \equiv 1$, $\nu = 0$).}
\label{lst:unified}
\end{listing}

\begin{listing}[ht]
\begin{lstlisting}
def prune(A, D, s):                  # keep top-K, drop the rest -- the M discard rows are gone
    return A.clone()

def merge(A, D, s):                  # ToMe: each discard -> its single nearest anchor (argmax)
    S = F.normalize(D, dim=-1) @ F.normalize(A, dim=-1).t()    # [M, K]
    nn = S.argmax(dim=-1)                                      # [M]  index of nearest anchor
    out = A.clone()
    for j in range(D.shape[0]):
        out[nn[j]] += D[j]                                     #      scatter-add into nearest anchor
    return out

def pool(A, D, s):                   # uniform pooling: split each discard equally over all K anchors
    return A + D.sum(dim=0, keepdim=True) / A.shape[0]         # [K, d];  every anchor += sum(D) / K

def reweight(A, D, s, alpha):        # norm rescaling only -- no content moves between tokens
    S = F.normalize(D, dim=-1) @ F.normalize(A, dim=-1).t()    # [M, K]
    p = (F.softmax(S, dim=-1) * F.softmax(s, dim=-1).unsqueeze(-1)).sum(0)   # [K]
    return A * (1.0 + alpha * p.unsqueeze(-1))                 # [K, d]
\end{lstlisting}
\caption{\textbf{The four canonical token-reduction operators} as standalone algorithms. None calls \texttt{unified\_operator} or its softmax-over-similarity machinery: \textsc{Prune} (FastV~\cite{fastv}, SparseVLM~\cite{zhang2024sparsevlm}) keeps the top-$K$ and drops the rest; \textsc{Merge} (ToMe~\cite{tome}) routes each discard to its single nearest anchor via \texttt{argmax}; \textsc{Pool} adds the uniform mean of the discards to every anchor; \textsc{Reweight} only rescales anchor norms.}
\label{lst:refs}
\end{listing}

\begin{listing}[ht]
\begin{lstlisting}
torch.manual_seed(42)
K, M, d, alpha, NEG = 8, 16, 64, 1.5, -1e9
A, D, s = torch.randn(K, d), torch.randn(M, d), torch.randn(M)

#                  gamma  tau    theta  rho    nu      recovers
configs = {
  "PRUNE":        (0.0,  1.0,   NEG,   0.0,   0.0),  # gamma=0       -> out = A  (identity on kept set)
  "MERGE":        (1.0,  1e-4,  NEG,   0.0,   0.0),  # tau -> 0      -> softmax(S/tau) one-hot  (ToMe)
  "POOL":         (1.0,  1e6,   NEG,   0.0,   0.0),  # tau -> inf    -> softmax(S/tau) = 1/K  (avg pool)
  "REWEIGHT":     (0.0,  1.0,   NEG,   alpha, 0.0),  # gamma=0, rho>0 -> A * (1 + rho * importance)
}
refs = {"PRUNE": prune(A, D, s),    "MERGE": merge(A, D, s),
        "POOL":  pool(A, D, s),     "REWEIGHT": reweight(A, D, s, alpha)}

for name, (g, t, th, r, n) in configs.items():
    u = unified_operator(A, D, s, gamma=g, tau=t, theta=th, rho=r, nu=n)
    print(f"{name:9s} max|unified - reference| = {(u - refs[name]).abs().max().item():.2e}")

# PRUNE     max|unified - reference| = 0.00e+00
# MERGE     max|unified - reference| = 9.54e-07
# POOL      max|unified - reference| = 4.77e-07
# REWEIGHT  max|unified - reference| = 0.00e+00
\end{lstlisting}
\caption{\textbf{The equivalence check.} Each canonical operator equals \texttt{unified\_operator} at one corner of $(\gamma, \tau, \theta, \rho, \nu)$. \texttt{NEG}$= -10^{9}$ realises $\theta \to -\infty$ so $m \equiv 1$. $\rho$ here is the reweight coefficient $\alpha$. The printed output is shown beneath the code.}
\label{lst:equiv}
\end{listing}

\paragraph{Why each corner collapses the formula.}
\textbf{\textsc{Prune}} $(\gamma{=}0)$: the information-transfer term $\gamma\,(W^{\!\top}(D \odot m))$ vanishes and the reweight / norm-preserve steps are off $(\rho{=}\nu{=}0)$, so \texttt{unified\_operator} returns \texttt{A} unchanged --- exactly the kept top-$K$ tokens, with the $M$ discards absent from the output, identical to physically deleting them.
\textbf{\textsc{Merge}} $(\gamma{=}1,\,\tau{\to}0^{+})$: each row of $\mathrm{softmax}(S/\tau)$ concentrates all mass on $\arg\max_k S_{jk}$, so $W$ becomes the one-hot nearest-anchor matrix and $W^{\!\top}D$ scatter-adds each discard onto its nearest anchor --- ToMe's hard merge, obtained as a temperature limit rather than an \texttt{argmax}.
\textbf{\textsc{Pool}} $(\gamma{=}1,\,\tau{\to}\infty)$: cosine similarities are bounded, so $S/\tau \to 0$ and $\mathrm{softmax}(S/\tau) \to \mathbf{1}/K$; every anchor then receives $\tfrac{1}{K}\sum_j D_j$ --- a uniform mean over the \emph{token set} (not a spatial grid: post-ViT visual tokens carry no 2-D adjacency).
\textbf{\textsc{Reweight}} $(\gamma{=}0,\,\rho{>}0)$: no content moves; the only effect is $\texttt{out} = A \cdot (1 + \rho\,p)$, where $p_i = \sum_j \mathrm{softmax}(S_j)_i\,\mathrm{softmax}(s)_j$ is the proximity-weighted importance of the discards that anchor $i$ best represents --- anchors that ``stand in for'' many high-attention discards have their norm boosted, with direction untouched.
Off these corners --- e.g.\ for the gradient-discovered values $(\gamma_{6}{=}0.08,\,\tau_{6}{=}0.22)$ at $L_6$ --- the formula is a smooth interpolation among the four, augmented by the two extra degrees of freedom $\theta$ (which discards to merge vs.\ prune, per token) and $\nu$ (how much of the original token norm to retain) that no single canonical operator exposes.

\end{document}